\documentclass[twoside]{article}

\usepackage[accepted]{aistats2022}
\makeatletter

\usepackage{microtype}
\usepackage{graphicx}
\usepackage[aboveskip=2pt]{subcaption}
\usepackage{standalone}
\usepackage{booktabs} 
\usepackage{tabularx}
\usepackage{mathtools}
\usepackage{amsmath,amsfonts,amssymb,amsthm}

\usepackage{tikz,pgfplots}
\usetikzlibrary{plotmarks,shapes,shapes.arrows,positioning}
\pgfplotsset{compat=newest} 
\pgfkeys{/pgf/number format/.cd,1000 sep={}}

\usepackage{commath}
\usepackage{enumitem}
\usepackage{systeme}
\usepackage{accents}
\usepackage{yfonts}
\usepackage{appendix}
\usepackage{svg}
\usepackage{bm}
\usepackage{xr}
\usepackage{pgfplots}
\usepackage{adjustbox}
\usepackage{dirtytalk}
\usepackage{natbib}
\usepackage[colorlinks=true,citecolor=black,urlcolor=black,linkcolor=black]{hyperref}
\usepackage{algorithm}
\usepackage{algorithmic}
\usepackage{mdframed}
\usepackage{wrapfig}

\usepackage[capitalize,nameinlink]{cleveref}
\crefname{section}{Sec.}{Sects.}
\crefname{proposition}{Prop.}{Props.}
\crefname{lemma}{Lem.}{Lems.}
\crefname{model}{Mod.}{Mods.}
\crefname{appendix}{App.}{Apps.}

\usepgfplotslibrary{groupplots}

\makeatletter
\newcommand*{\addFileDependency}[1]{
  \typeout{(#1)}
  \@addtofilelist{#1}
  \IfFileExists{#1}{}{\typeout{No file #1.}}
}
\makeatother

\graphicspath{ {images/} }

\newtheorem{definition}{Definition}[section]

\newtheorem{theorem}{Theorem}
\newtheorem{corollary}{Corollary}[theorem]
\newtheorem{lemma}[theorem]{Lemma}
\newtheorem{proposition}{Proposition}
\newenvironment{FrameProposition}
  {\begin{mdframed}[linecolor=black!5,backgroundcolor=black!5,roundcorner=3pt,innerleftmargin=4pt,innerrightmargin=4pt,innertopmargin=6pt,innerbottommargin=6pt]\begin{proposition}}
  {\end{proposition}\end{mdframed}}

\newcommand*{\dt}[1]{%
  \accentset{\mbox{\small\bfseries .}}{#1}}
\newcommand*{\ddt}[1]{%
  \accentset{\mbox{\small\bfseries .\hspace{-0.10ex}.}}{#1}}

\DeclareMathOperator{\EX}{\mathbb{E}}

\setlist{topsep=0pt}

\let\oldtextbf\textbf
\renewcommand{\textbf}[1]{\oldtextbf{\boldmath #1}}

\newcommand{\Cov}{\mathrm{Cov}}
\newcommand*{\fractionalLaplacian}{\bm{\widetilde{L}}}
\newcommand*{\transposed}{^\intercal}

\newcommand{\nipstitle}[1]{{%
    \phantomsection\hsize\textwidth\linewidth\hsize%
    \vskip 0.1in%
    \toptitlebar%
    \begin{minipage}{\textwidth}%
        \centering{\Large\bf #1\par}%
    \end{minipage}%
    \bottomtitlebar%
    \addcontentsline{toc}{section}{#1}%
}}

\begin{document}

\runningauthor{Alexander Nikitin, ST John, Arno Solin, Samuel Kaski} 

\twocolumn[

\aistatstitle{Non-separable Spatio-temporal Graph Kernels via SPDEs}
\aistatsauthor{{Alexander Nikitin}\textsuperscript{\ensuremath{1}} \And {ST John}\textsuperscript{\ensuremath{1}} \And {Arno Solin}\textsuperscript{\ensuremath{1}} \And {Samuel Kaski}\textsuperscript{\ensuremath{1, 2}}}
\aistatsauthor{}
\aistatsaddress{\textsuperscript{\ensuremath{1}}Finnish Center for Artificial Intelligence FCAI, Department of Computer Science, Aalto University, Finland \\ \textsuperscript{\ensuremath{2}}Department of Computer Science, University of Manchester, UK}]

\begin{abstract}
    Gaussian processes (GPs) provide a principled and direct approach for inference and learning on graphs. However, the lack of justified graph kernels for spatio-temporal modelling has held back their use in graph problems. We leverage an explicit link between stochastic partial differential equations (SPDEs) and GPs on graphs, introduce a framework for deriving graph kernels via SPDEs, and derive non-separable spatio-temporal graph kernels that capture interaction across space and time. We formulate the graph kernels for the stochastic heat equation and wave equation. We show that by providing novel tools for spatio-temporal GP modelling on graphs, we outperform pre-existing graph kernels in real-world applications that feature diffusion, oscillation, and other complicated interactions. 
\end{abstract}

\section{INTRODUCTION}

Contemporary machine learning typically leverages structure within the modelling task or application. Often, the structure can be represented as a graph, where the objects and relations are represented as nodes and edges. Such tasks can be found, for example, within chemoinformatics \citep{chemoinformatics}, social network analysis \citep{social_network_analysis}, and anomaly detection \citep{akoglu2015graph}. The abundance of graph-structured problems has led to the invention of new methods for graph-structured data, including graph neural networks \citep[GNNs,][]{scarselli2008graph} and spectral methods \citep{chung1997spectral} for capturing structure across space and time.

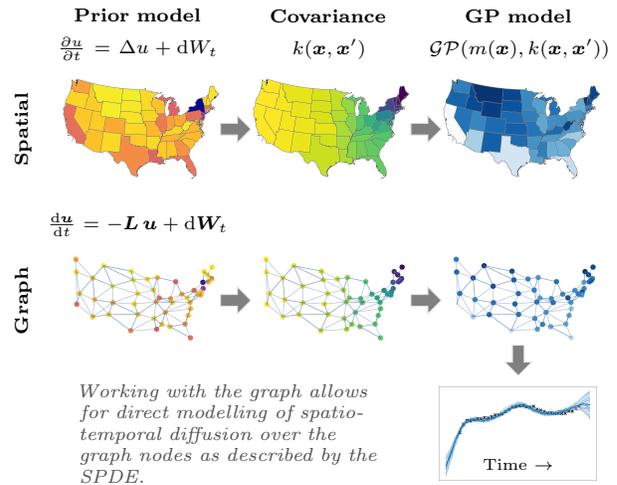
\begin{figure}[!t]
  \centering\scriptsize
  \resizebox{\columnwidth}{!}{%
  \begin{tikzpicture}[outer sep=0]
  
    \node[] at (-2, 1.5) {\includegraphics[width=.33\columnwidth]{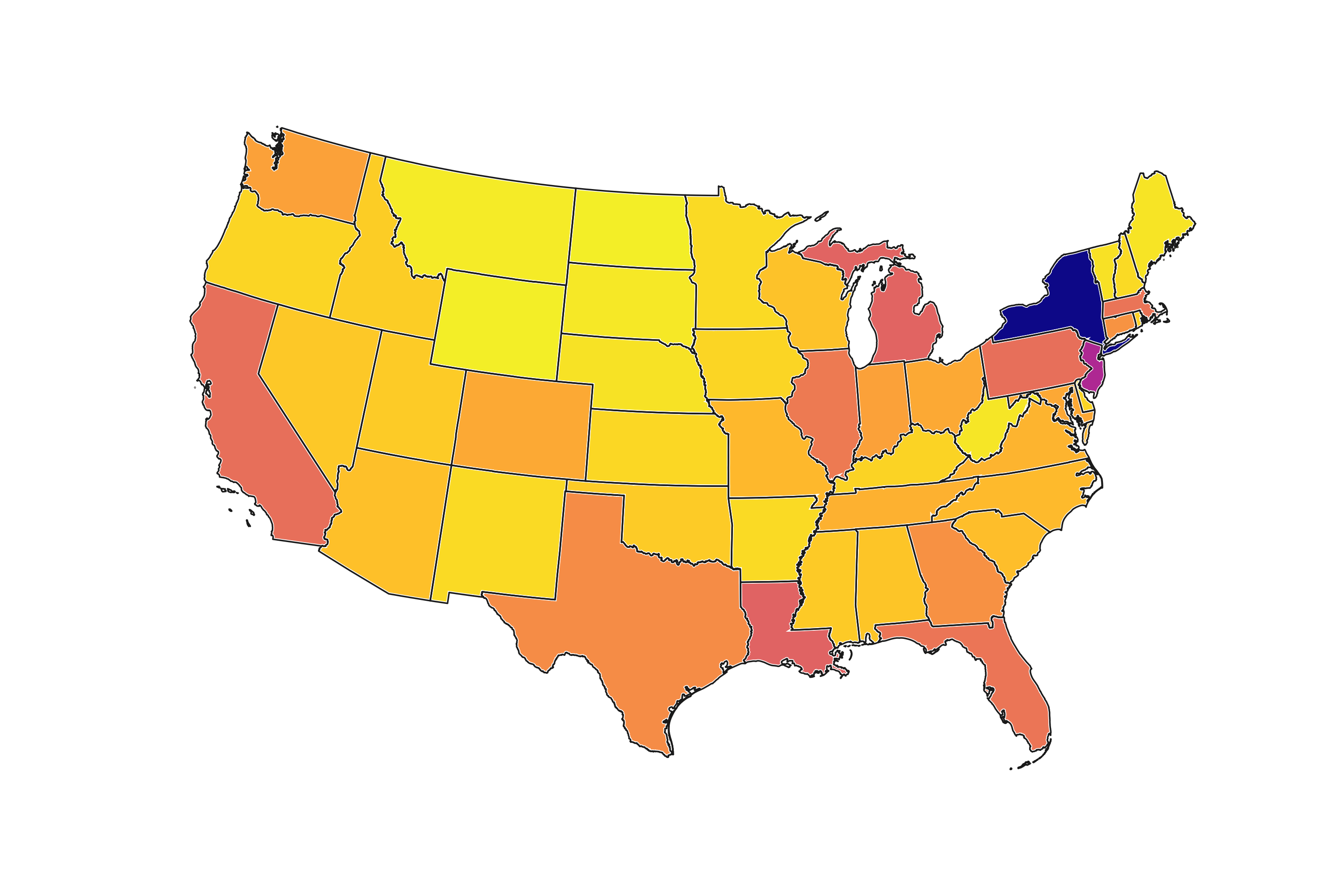}};
    
    \node[] at (0.5, 1.5) {\includegraphics[width=.33\columnwidth]{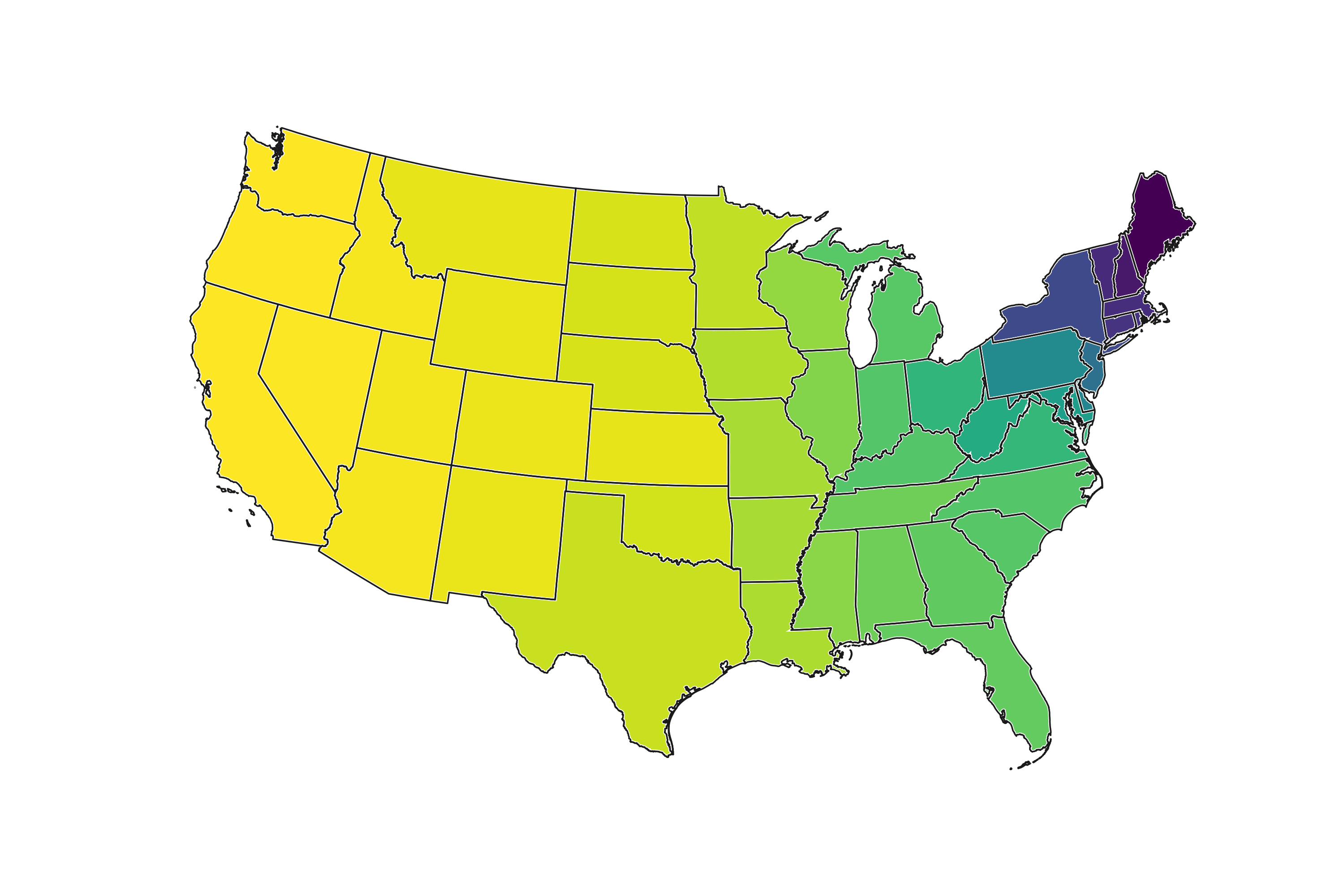}};
    
    \node[] at (3, 1.5) {\includegraphics[width=.33\columnwidth]{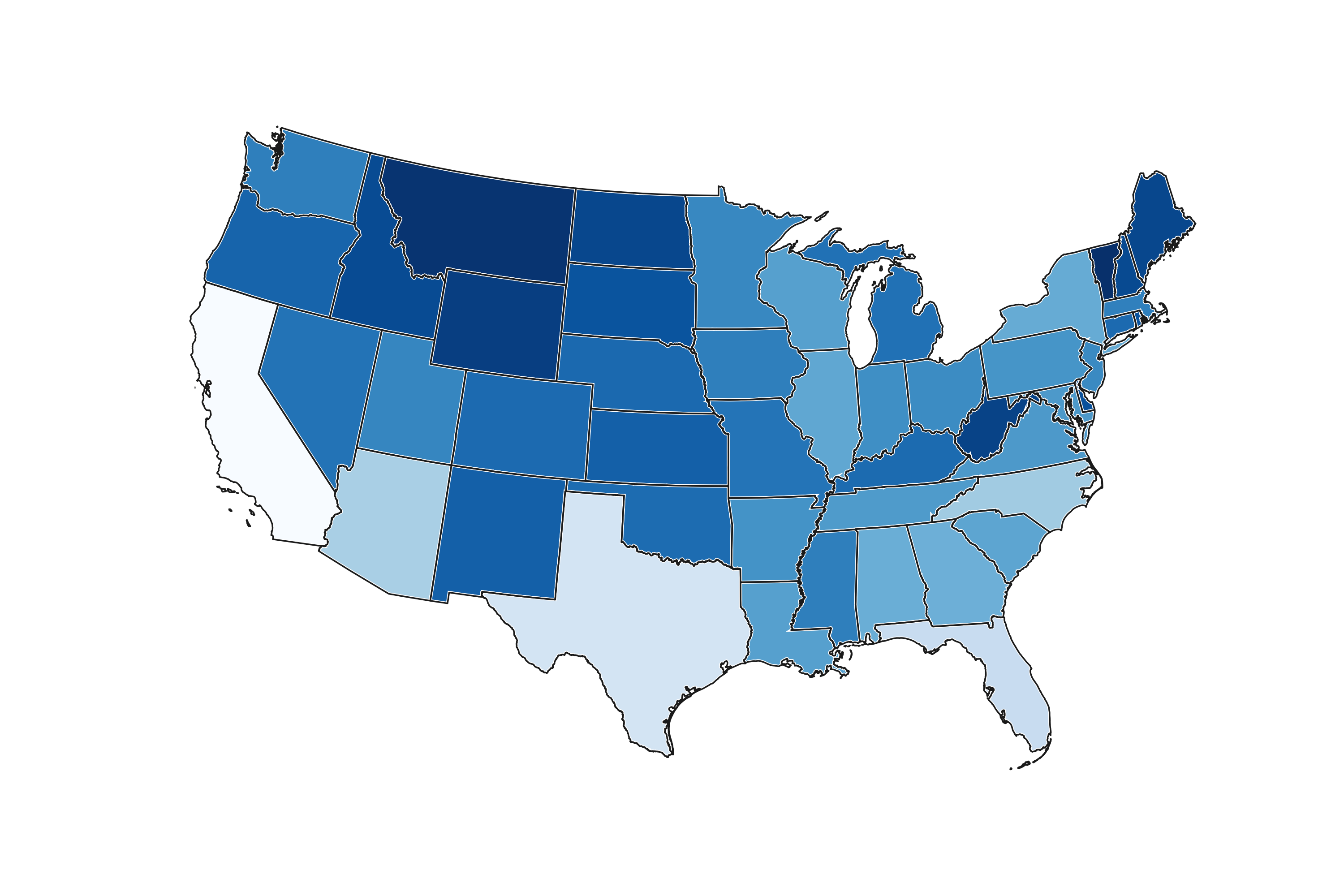}};
    
    \node[] at (-2, -0.75) {\includegraphics[width=.33\columnwidth]{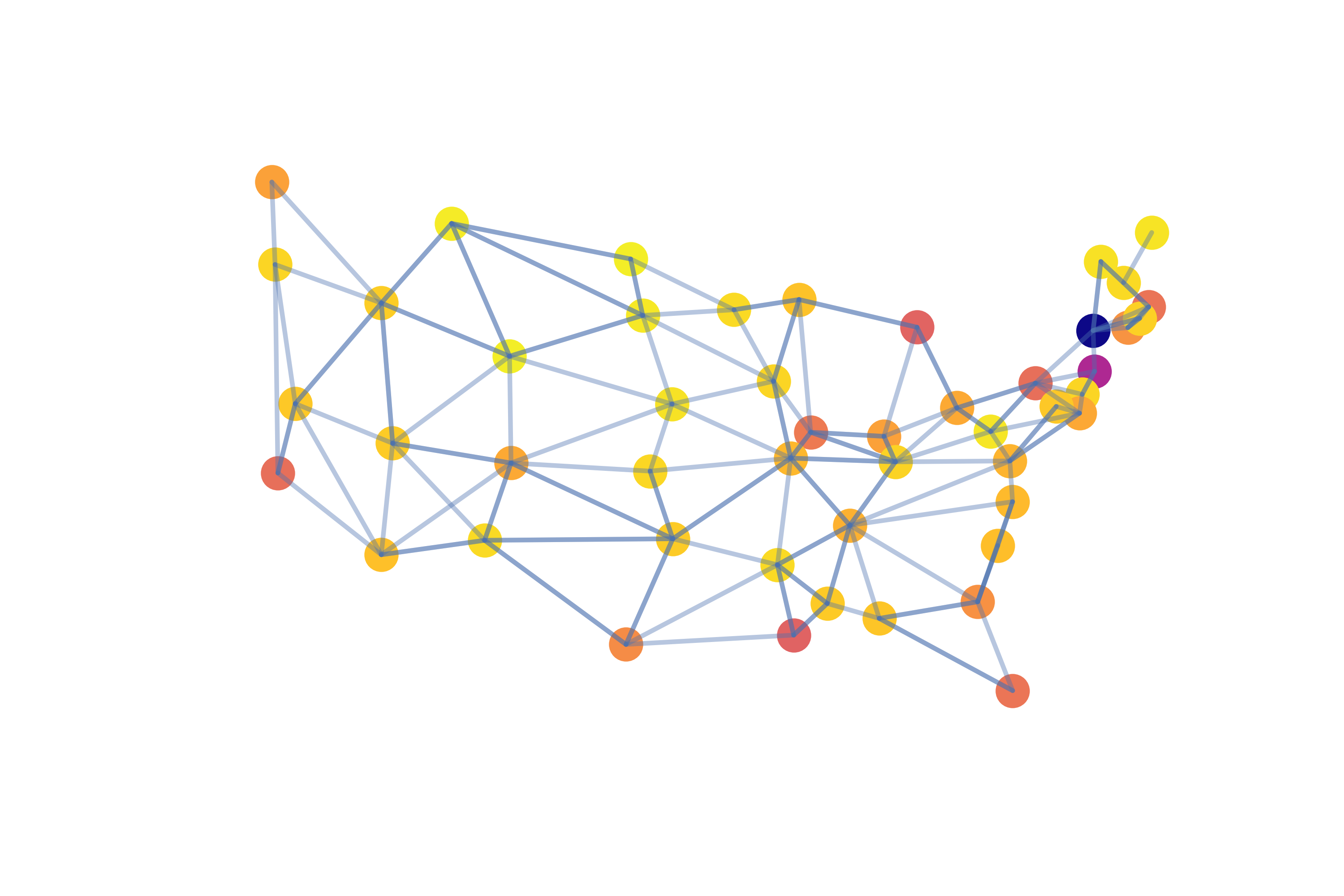}};
    
    \node[] at (0.5, -0.75) {\includegraphics[width=.33\columnwidth]{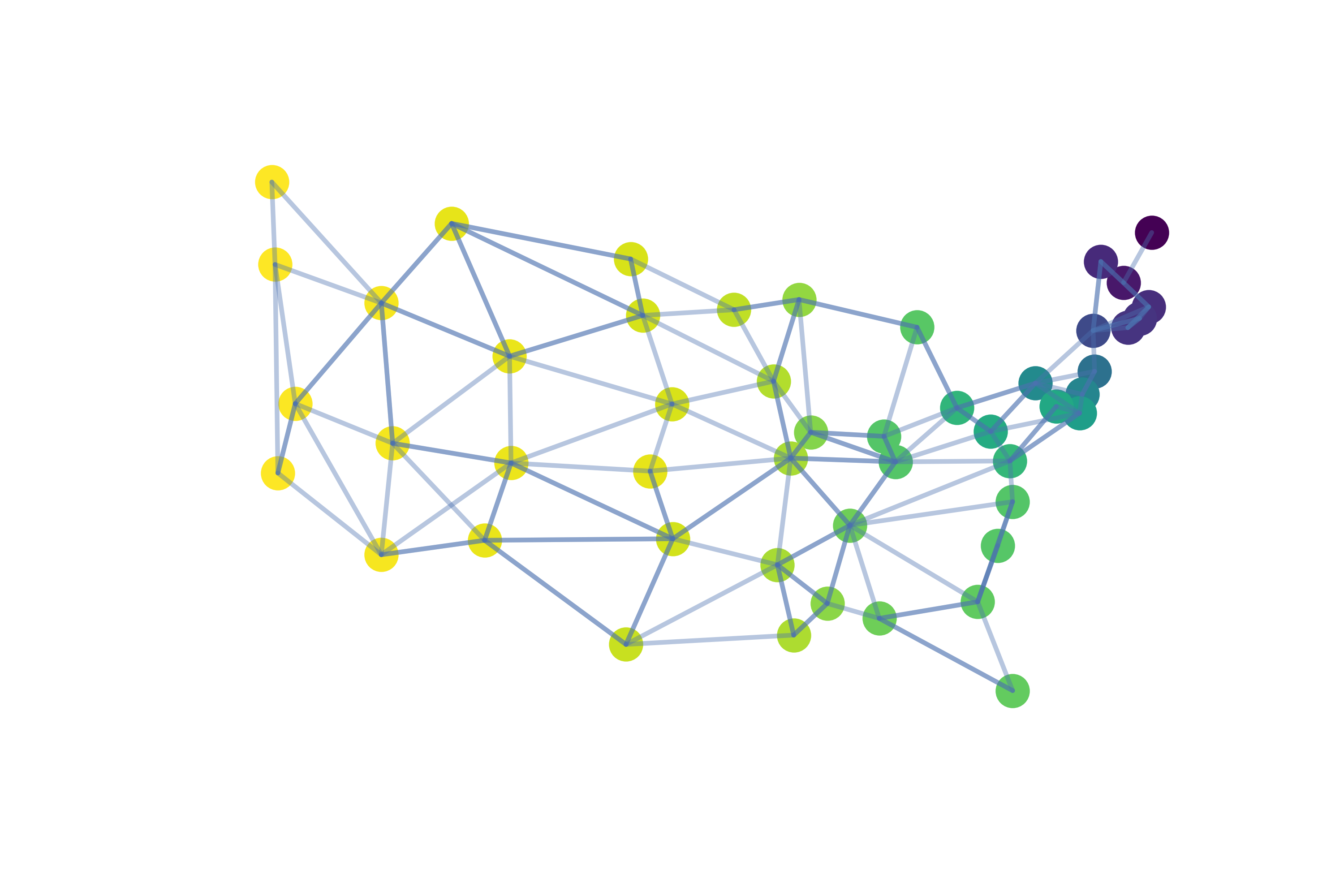}};
    
    \node[] at (3, -0.75) {\includegraphics[width=.33\columnwidth]{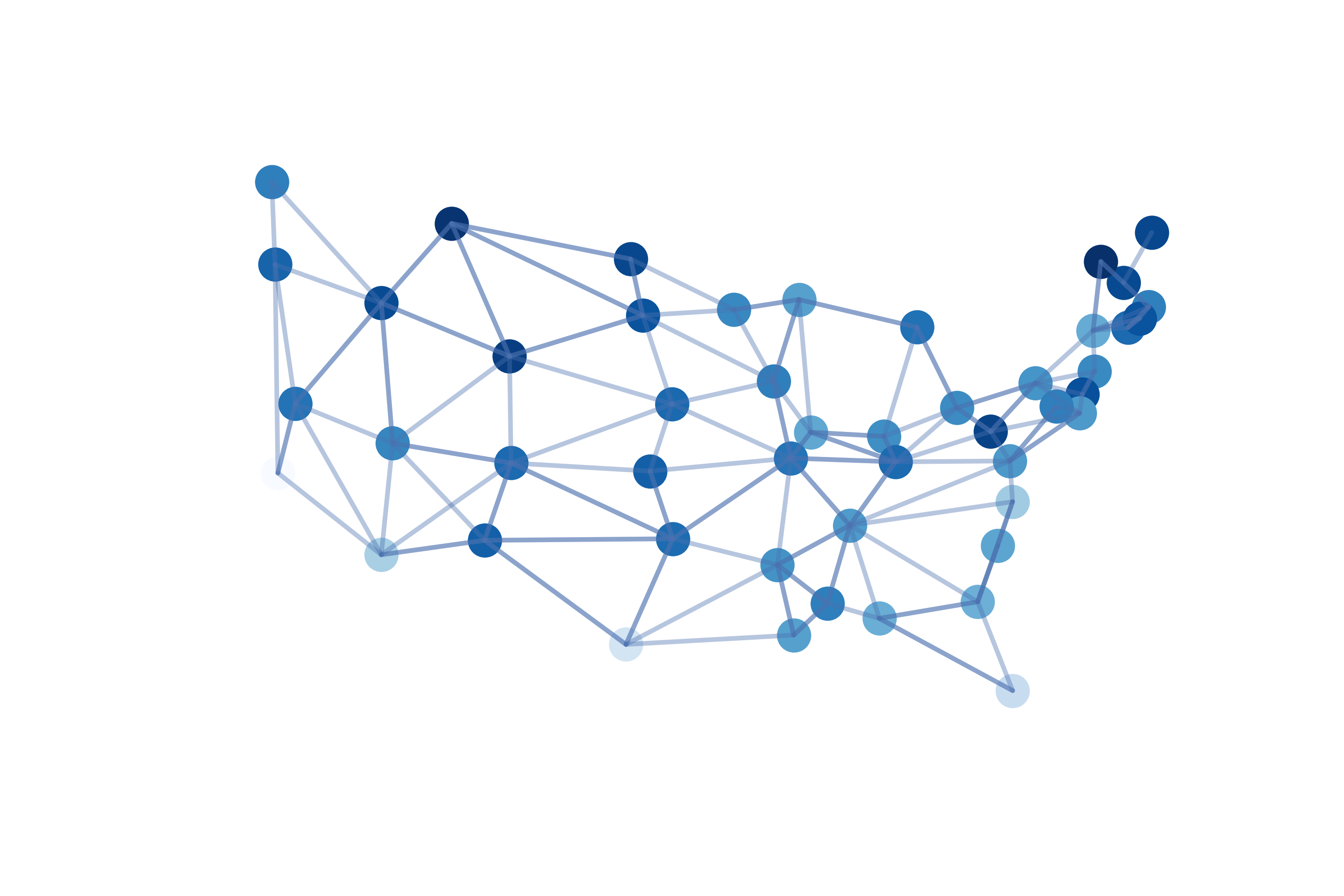}};
    
    \node[] at (3, -2.5) {\includegraphics[width=.25\columnwidth]{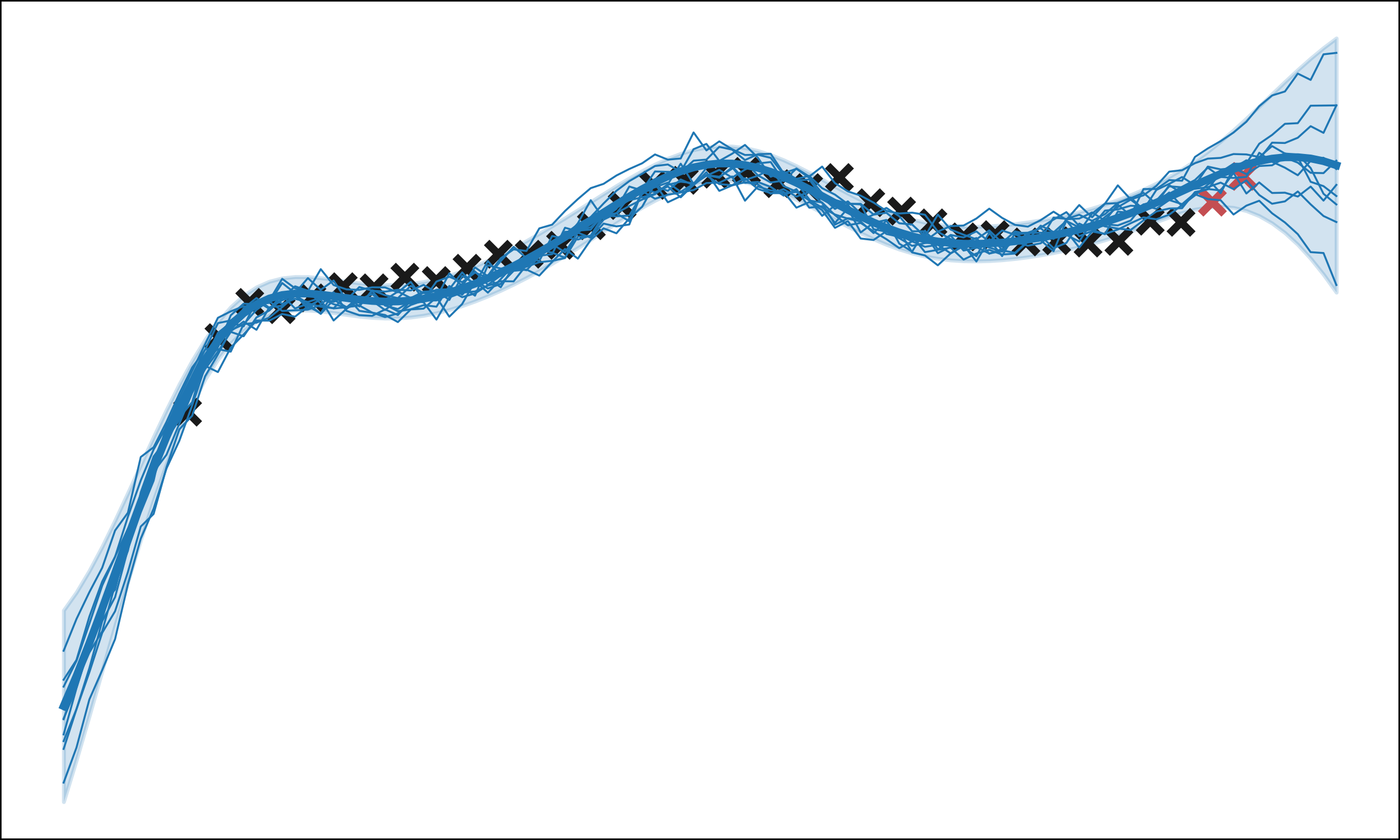}};
    
    \tikzstyle{arrow} = [draw=black!10, single arrow, minimum height=4mm, minimum width=2mm, single arrow head extend=1mm, fill=black!50, anchor=center, rotate=0, inner sep=2pt]    
     \node[arrow] at (-0.75,1.5) {};
     \node[arrow] at (1.75,1.5) {};
     \node[arrow] at (-0.75, -0.75) {};
     \node[arrow] at (1.75, -0.75) {};
     
     \node[arrow,rotate=-90] at (3, -1.5) {};

     \tikzstyle{label}=[text width=2.5cm, align=center]
     \node[label] at (-2,2.75) {\bf Prior model \\[4pt] $\frac{\partial u}{\partial t} = \Delta u + \mathrm{d}W_t$};
     \node[label] at (0.5, 2.75) {\bf Covariance\\[4pt] $k(\bm{x},\bm{x}')$};    
     \node[label] at (3, 2.75) {\bf GP model \\[4pt] $\mathcal{GP}(m(\bm{x}),k(\bm{x}, \bm{x}'))$}; 
    
     \node[label,rotate=90] at (-3.5, 1.5) {\bf Spatial};
     \node[label,rotate=90] at (-3.5, -0.75) {\bf Graph};
     
     \node[label] at (-2, 0.25) {\bf $\frac{\mathrm{d}\bm{u}}{\mathrm{d}t} = -\bm{L}\,\bm{u} + \mathrm{d}\bm{W}_t$};
     
     \node at (3, -2.9) {\tiny Time $\rightarrow$};
     
     \node[text width=4cm] at (-0.75, -2.5) {\color{black!70}\it Working with the graph allows for direct modelling of spatio-temporal diffusion over the graph nodes as described by the SPDE.};
                
  \end{tikzpicture}}\\
  \caption{We start with an unknown process on a particular domain with an SPDE model, discretize it, and derive the GP covariance function from this SPDE. Finally, we use GP inference in the spatial or spatio-temporal model from discrete graph-structured observations.}
  \label{fig:teaser}
\end{figure}

In this work, we focus on graph-based prediction of temporal signals on graphs. In other words, the goal is to approximate spatio-temporal processes on graphs. Many real-life problems belong to this class of tasks, including traffic prediction \citep{jiang2021graph}, epidemiological modelling \citep{keeling2005networks}, and analysis of dynamics in social networks \citep{carrington2005models}. Recently, a number of approaches have been developed for temporal machine learning problems on graphs, including the works by \citet{li2017diffusion}, \citet{yu2017spatio}, and \citet{zhao2019t}.

We approach these graph-based learning problems using Gaussian processes \citep[GPs,][]{Rasmussen+Williams:2006}. GPs are widely used tools in statistical machine learning, where including prior knowledge is desired. Their probabilistic treatment allows for convenient inference machinery and enables capturing the uncertainties associated with predictions. However, it is not always apparent how to encode relevant prior knowledge into the GP prior on graphs. In the continuous domain, bringing in prior knowledge from stochastic partial differential equations (SPDEs) has enabled constructing useful kernels, including the Mat\'ern family of kernels \citep{whittle1963stochastic, lindgren2011explicit}, resulting in improvements in quality and performance in many tasks. In this work, we ask whether analogous constructs can be made for graphs and whether they improve the performance of GPs on spatio-temporal graph problems. We hypothesize that because many physical processes follow differential equations, deriving kernels from their stochastic counterparts will allow incorporating physical priors in GP models.

Apart from challenges in the model specification for graph-based GPs, the usual drawbacks also apply: closed-form inference is only possible for conjugate observation models, and a na\"ive treatment results in a cubic computational scaling with respect to the number of observations \citep{Rasmussen+Williams:2006}. This means that vanilla GPs are not particularly suitable for large-scale problems. However, a number of modern methods exist to overcome this issue, including sparse Gaussian processes using inducing points \citep[see][for a recent review]{liu2020gaussian}.
The development of SPDE-based approaches to Gaussian processes has led to several practical solutions. One of those is a popular R package for approximate Bayesian inference: R-INLA \citep{lindgren2015bayesian} that uses SPDEs to sample from spatial models. The approach we propose in this paper allows extending this framework to graph problems, and it is illustrated in \cref{fig:teaser}.

\textbf{The contributions of this paper:} {\em (i)}~We extend methods from spatial statistics and, hence, link GPs to SPDEs for temporal signals on graphs. 
{\em (ii)}~We apply this approach to spatio-temporal modelling on graphs and consider the analogs of well-known SPDEs from spatial statistics, which allows us to derive non-separable spatio-temporal kernels on graphs.
{\em (iii)}~We empirically evaluate these kernels and show their effectiveness on both synthetic data sets and applied machine learning problems: prediction of the distribution of chickenpox and COVID-19 epidemic.

\section{RELATED WORK}
\label{section:related_work}

The connection between Gaussian processes and stochastic (partial) differential equations was established by \citet{whittle1963stochastic}. It resulted in a number of works in spatial statistics, most notably the work by \citet{lindgren2011explicit}, the library R-INLA \citep{lindgren2015bayesian}, and by \citet{solin2016stochastic} in the spatio-temporal context. These methods have been applied in many domains, for example, animal movement prediction \citep{hooten2017animal}, geographical information system modelling \citep{burrough2015principles}, and geostatistical analysis \citep{moraga2017geostatistical}.

To apply SPDE approaches to graphs, we need to introduce spectral graph theory, which has a long history of research and applications \citep[e.g.,][]{chung1997spectral,von2007tutorial}. These works typically consider the spectrum of the graph Laplacian and derive properties that can be helpful for different tasks on graphs. \citet{belkin2003laplacian} used spectral graph theory to develop invariant embedding maps. These embedding maps then showed good results in dimensionality reduction \citep{graph_embedding_and_extensions}. Spectral graph theory is linked to GPs on graphs by considering suitable spatial (pseudo-)differential operators and their associated kernels (covariance functions). 

A number of graph kernels were developed using the methods from spectral graph theory. \citet{smola2003kernels} proposed the construction of kernels on graphs by introducing random walk and heat kernels on graphs, evaluated their performance, and derived necessary properties of the kernels. Recent work on graph approaches to Mat\'ern fields considers graph discretization and theoretically compares the covariances produced by discrete approximation and Mat\'ern kernels \citep{sanz2020spde}. Graph representation learning has gained popularity because of its large number of applications \citep{chami2020machine}. Kernel methods, such as kernel ridge regression, for prediction on graphs were proposed by \citet{romero2016kernel}. This approach was extended to spatio-temporal signals on dynamic graphs \citep{romero2017kernel}.
Mat\'ern graph kernels were presented by recent works: Mat\'ern kernels on manifolds \citep{matern_on_manifolds}, and, then, on graphs \citep{matern_on_graphs}.

We extend and generalize this approach with the SPDE framework for graph kernels and then go beyond spatial graph kernels to the spatio-temporal domain.

\section{BACKGROUND}
\label{section:background}
Gaussian processes are a non-parametric machine learning paradigm, in which we model the target function as a stochastic process, whose evaluation at any finite set of points has a joint Gaussian distribution \citep{Rasmussen+Williams:2006}. A Gaussian process $f(\bm{x}) \sim \mathcal{GP}(m(\bm{x}), k(\bm{x}, \bm{x}^{\prime}))$ is defined by its mean function $m(\bm{x})$ and covariance function $k(\bm{x}, \bm{x}^{\prime})$ \citep{bishop2006pattern}. The covariance function is often called a `kernel'. The kernel encapsulates prior knowledge, and defining a good kernel is one of the key ingredients and challenges of setting up the GP model \citep{duvenaud2014automatic}. GPs can be extended to vector-valued functions using multioutput GPs \citep{MOGP}. Kernels then become matrix-valued: $\bm{K}(\bm{x}, \bm{x}^{\prime}): \mathbb{R}^{n} \times \mathbb{R}^{n} \rightarrow \mathbb{R}^{m \times m}$, for $n$-dimensional inputs and $m$-dimensional outputs. Standard GP toolchains include various kernels on continuous domains (e.g., periodic, polynomial, and Mat\'ern family). However, application of GPs to other domains is often restricted by the lack of principled kernels.

\subsection{Stochastic Partial Differential Equations}
In spatial statistics, Gaussian random fields can be represented as solutions to an SPDE. The physical sense of an underlying process can be included in GPs in a natural way by forming the prior covariance function as a solution to an SPDE \citep{sarkka_spde}. Many widely used GP kernels can be derived from the corresponding SPDEs. For example, the Mat\'ern kernel family for multidimensional GPs can be derived from underlying SPDEs \citep{whittle1963stochastic}.

Stochastic partial differential equations (SPDEs) generalize partial differential equations via introducing random forces (noise) to some terms and coefficients, similarly to how stochastic differential equations generalize ordinary differential equations \citep{oksendal2013stochastic, gardiner2004handbook, sarkka2019applied}. SPDEs are applied in various fields, including physics, signal processing, and machine learning \citep{schottl1997stochastic}.

\subsection{Spectral Graph Theory}
Spectral graph theory allows us to derive graph properties by operating with the spectrum of its Laplacian matrix (a.k.a.\ graph Laplacian), which will allow us to define partial differential equations (PDEs) on graphs.
\begin{definition} The graph Laplacian $\bm{L}$ of a graph $G=\left(V, E\right)$ is the matrix
\begin{equation} \label{equation:weighted_laplacian_definition}
    \bm{L} = \bm{D}_{W} - \bm{W},
\end{equation}
where $\bm{W}$ is the matrix of the edge weights and $\bm{D}_{W}=\operatorname{diag}(w_i)$ is the diagonal matrix of the accumulated weights $w_i = \sum\limits_{j: (i, j) \in E} w_{ij}$. In the unweighted case, $\bm{D}_{W}$ is the degree matrix and $\bm{W}$ is the adjacency matrix. 
\end{definition}

The graph Laplacian can be normalized to have unit diagonal entries. All derivations in this article apply to both normalized and non-normalized Laplacians in weighted and unweighted graphs. In practice, the choice between normalized and unnormalized Laplacians depends on the graph structure and should be based on the particular modeling task.

\section{SPATIO-TEMPORAL KERNELS ON GRAPHS VIA SPDEs}
\label{section:theory}
We consider a spatio-temporal prediction problem on a graph $G=(V, E)$ given a data set $\left\{\left(x_i, y_i\right)\right\}_{i=1}^{N}$, with features $x_i \in D$, where the domain $D = V \times \mathbb{R}$ describes graph vertex and time. We focus on the regression task, where $y_i$ are the targets from the real numbers $\mathbb{R}$. We aim to approximate a function $f$ such that $f: D \to \mathbb{R}$. The methods we propose are applicable for classification problems $f: D \to C$ for a finite set of labels $C$ as well.

\textbf{Notation.} If not stated otherwise, we denote
$f(\bm{x}): \mathbb{R}^{n} \to \mathbb{R}$ a continuous spatial function,
$\bm{v} \in \mathbb{R}^{|V|}$ a spatial function on a graph (i.e., a vector whose components are the function values at the vertices),
and $\bm{u}(t): \mathbb{R} \to \mathbb{R}^{|V|}$ a spatio-temporal function on a graph.
We will denote $w(\bm{x})$ spatial white noise%
\footnote{I.e., $\mathbb{E}[w(\bm{x})]=0$, $\operatorname{Cov}[w(\bm{x}),w(\bm{x}')]=\delta(\bm{x}-\bm{x}')$.}%
,
$\bm{w}$ a $|V|$-dimensional white noise%
\footnote{I.e., $ \mathbb{E}[\bm{w}] = \bm{0}$ and $\mathbb{E}[\bm{w} \bm{w}\transposed] = \bm{I}$.}%
,
and $\mathrm{d}\bm{W}_t$ the differential of $|V|$-dimensional Brownian motion. $\dt{\bm{W}_t} = \frac{\mathrm{d}\bm{W}_t}{\mathrm{d} t}$ is the \emph{formal} temporal derivative of Brownian motion\footnote{Even though Brownian motion is nowhere differentiable, assuming its derivative to be temporal white noise is the standard way of treating it in SDE literature as discussed by \citet{sarkka2019applied}.}, and $\bm{W}_t$ is Brownian motion.

\subsection{SPDE Framework for Graph Kernels}
\label{section:spde_framework}
Our framework for deriving graph kernels via SPDEs consists of the following high-level steps: {\em {(i)}} define an SPDE using prior knowledge about the underlying process, { \em{(ii)}} convert it to a graph counterpart, { \em{(iii)}} solve the graph counterpart, and { \em{(iv)}} derive corresponding mean and covariance function of GP on graph. Next, we will provide specific examples and use this framework to derive novel kernels.

Many PDEs are defined using the Laplace operator $\Delta$.
In Euclidean space, it is defined as $\Delta = \sum_{i=1}^{n}\frac{\partial^2}{\partial x_i^2}$. On graphs, we define as the corresponding operator the negative graph Laplacian $-\bm{L}$. This is justified by considering a lattice approximation of the real plane and ensuring that this operator represents exactly the discretization of $\Delta$ \citep{wardetzky2020discrete}.
    
We illustrate the approach by considering the analogue of Laplace's equation on graphs. Laplace's equation, $\Delta f(\bm{x}) = 0$, is a second-order partial differential equation relevant across physics, for example, in fluid dynamics and electromagnetism.
Its stochastic counterpart, known as Laplace's stochastic partial differential equation, is $\Delta f(\bm{x}) = w(\bm{x})$.
For the analogue of Laplace's SPDE on a graph, we replace $\Delta$ with $-\bm{L}$ and arrive at
\begin{equation}
    -\bm{L} \bm{v} = \bm{w}.
    \label{eq:laplace-spde-graph}
\end{equation}

Because the space we are considering is discrete and the graph Laplacian replaces the Laplace operator, the graph counterparts of SPDEs become multi-dimensional SDEs.

For a formal derivation using discrete calculus, one can consider $\bm{v}$ as a 0-cochain, as in \citet{grady2010discrete}. Within our work, it is enough to operate with $\bm{v}$ as a vector. \cref{eq:laplace-spde-graph} defines the notion of harmonic functions on graphs: in functions that are solutions to \cref{eq:laplace-spde-graph}, the value at each node is given by the mean of adjacent nodes.

The solution $\bm{v}$ of \cref{eq:laplace-spde-graph} can be described as a Gaussian process (which in the discrete case, by definition, is a multivariate normal distribution): 
\begin{equation} \label{equation:laplacian_kernel}
    \bm{v} \sim \mathcal{N}(\bm{0}, (\bm{L}^{\top} \bm{L})^{+}) ,
\end{equation}
where $\bm{A}^{+}$ denotes the Moore--Penrose pseudoinverse of matrix $\bm{A}$ \citep{moore1920reciprocal}. This defines the Laplacian kernel on graphs, $\bm{K} = (\bm{L}^{\top} \bm{L})^{+}$.

Another example of the SPDE framework applied to graph kernels is the Mat\'ern family of kernels. The Mat\'ern kernel family is an essential kernel family for many applications of GPs. The isotropic (rotation and translation invariant) Mat\'ern kernels are defined as \citep{Rasmussen+Williams:2006}
\begin{equation}
  k_{\text{Mat\'ern}}(r)=2^{1-\nu} \gamma^\nu K_\nu(\gamma)/\Gamma(\nu), \quad \gamma = \sqrt{2 \nu} r/\kappa,
\end{equation}
where $r=\|\bm{x}-\bm{x}'\|$ is the Euclidean distance, $\kappa$ is the lengthscale, and $\nu$ controls smoothness (samples from a GP with this kernel are $(\lceil\nu\rceil-1)$-times differentiable).
The Mat\'ern kernel in $\mathbb{R}^d$ can be derived from this SPDE with the fractional Laplace operator \citep{lindgren2011explicit}:
\begin{equation}
    \label{equation:matern_d_dimensional}
    \bigg(\frac{2 \nu}{\kappa^{2}}-\Delta\bigg)^{\frac{\nu}{2}+\frac{d}{4}} f(\bm{x}) = w(\bm{x}).
\end{equation}
This motivates us to use the fractional graph Laplacian:
\begin{definition} Fractional graph Laplacian $\widetilde{\bm{L}}$: 
\begin{equation}
    \widetilde{\bm{L}} = \bigg(\frac{2 \nu}{\kappa^{2}} \bm{I} + \bm{L}\bigg)^{\frac{\nu}{2}},
\end{equation}
where $\nu$ and $\kappa$ are positive scalar parameters.
\end{definition}
The fractional graph Laplacian is more flexible than $\bm{L}$ (for $\nu = 2$ and $\kappa \rightarrow \infty$, it coincides with $\bm{L}$). Additionally, for some combinations of hyperparameters, the fractional graph Laplacian possesses non-local properties (see \citet{benzi2020non}).

We can formulate \cref{equation:matern_d_dimensional} on graphs as
\begin{equation} \label{equation:matern_kernel_equation}
    \bm{\widetilde{L}} \bm{v}=\bm{w}.
\end{equation}
For a self-adjoint Laplacian $\bm{L}$ the covariance matrix is
\begin{equation}
\bm{K} = \bigg(\frac{2\nu}{\kappa^2} \bm{I} + \bm{L}\bigg)^{-\nu},
\end{equation}
and the solution of \cref{equation:matern_kernel_equation} is
$
    \bm{v} \sim \mathcal{N}\left(\bm{0}, \bm{K}\right)
$.
These results were recently considered by \citet{matern_on_graphs}, and here we showed how they fit into a broader SPDE framework.

\subsection{Separable Spatio-Temporal Kernels on Graphs}
\label{section:separable_kernels}
To obtain a kernel on the combination of two domains (here, graph and time), one can take kernels for each domain and combine them, commonly as a product: $k(\bm{x}, t; \bm{x}^{\prime}, t^{\prime}) = k_{\bm{x}}(\bm{x}, \bm{x}^{\prime}) \, k_t(t, t^{\prime})$, where $\bm{x}$ and $\bm{x}^{\prime}$ belong to the spatial domain and $t$ and $t^{\prime}$ to the temporal domain. The separability of a kernel is a choice made for convenience. An underlying process is not necessarily separable; an example of spatial covariance structure changing over time can be easily constructed: for instance, in spatio-temporal epidemiological modelling, the structure of the spatial covariance changes over time (e.g., travel restriction or development of a vaccine), thus it is not enough to separate spatial and temporal structure, and they have to be considered jointly.

\subsection{Non-Separable Kernels}
\label{section:non_separable_kernels}
We describe non-separable spatio-temporal graph kernels using the SPDE framework.
Let us consider two widely used parabolic and hyperbolic equations:
\begin{itemize}
    \item Heat equation: $\frac{\partial f}{\partial t} = c \Delta f$,
    \item Wave equation: $\frac{\partial^2 f}{\partial t^2} = c^2 \Delta f$.
\end{itemize}
We will define the stochastic counterparts of these equations on graphs, derive the corresponding covariance functions, and study them.
The kernel derivations from these two SPDEs are formulated as propositions. The full derivations are in Appendices~\ref{section:appendix_graph_kernels} and~\ref{section:appendix_wave_and_stochastic_heat}.

\subsection{Stochastic Heat Equation}
\label{section:stochastic_heat_equation}
The heat equation is written as $\frac{\partial f}{\partial t} = c \Delta{f}$. This PDE is called the \emph{heat} equation because it is used in physics for modeling heat transfer on surfaces. If $f$ is temperature, the Laplace operator on the equation's right-hand side determines the difference between the temperature at a certain point and its neighboring points on a surface.

We consider the discretization of this equation using the graph Laplacian. Let $\bm{u}(t)$ be a temporal signal on a graph which obeys the heat equation
\begin{equation} \label{equation:heat_equation}
  \frac{\mathrm{d} \bm{u}}{\mathrm{d} t} = -c \bm{L} \bm{u},  
\end{equation}
which is now an ODE with the solution
\begin{equation}
\begin{split}
    \bm{u}(t) = e^{-c \bm{L} t} \bm{u}(0).
\end{split}
\end{equation}
This solution is known as the heat kernel.
This result plays an important role in graph theory. The positive semi-definite matrix $e^{-c\bm{L}t}$ in this equation can be used as a kernel on the graph. This kernel is related to graph random walks, which are used widely in graph algorithms \citep{lovasz1993random}. To see this, consider a Taylor expansion of this kernel (assume $c=1$ for simplicity): $e^{-\bm{L} t} = \sum_{k=0}^{+\infty} \frac{{t^k} e^{-t}}{k!} \bm{P}^{k}$, where $\bm{P} = \bm{D}^{-1} \bm{A}$ is the random walk matrix of the graph. In fact, this kernel defines continuous-time random walks, where each jump occurs in $\mathrm{Poisson}(1)$ time, similarly to the continuous case.

By adding temporal white noise, we can construct a heat SPDE on graphs:
\begin{equation} \label{equation:stochastic_heat_equation_on_graph}
\frac{\mathrm{d} \bm{u}}{\mathrm{d} t} = -c \bm{L} \bm{u} + \sigma \dt{\bm{W}}_t.
\end{equation}

\begin{FrameProposition}
\label{theorem:stochastic_heat_kernel}
The \textbf{stochastic heat equation kernel (SHEK)} on graphs can be defined by adding spatio-temporal white noise, or for convenient integration, as a \emph{formal} derivative of the Wiener process $\dt{\bm{W}}_t$:
\begin{equation}
    \frac{\mathrm{d} \bm{u}}{\mathrm{d} t} = -c \bm{\widetilde{L}} \bm{u} + \sigma \dt{\bm{W}}_t.
\end{equation}
The solution is given by a Gaussian process:
\begin{flalign}
& \bm{u}(t) \sim \mathcal{GP}(\bm{\mu}(t), \Cov [\bm{u}(s), \bm{u}(t)]), \quad \text{with} \notag \\
   & \bm{\mu}(t) = e^{-c\bm{\widetilde{L}} t} \bm{u}(0), \notag\\
   & \Cov[\bm{u}(t), \bm{u}(s)] = \frac{\sigma^2}{c} e^{-c\bm{\widetilde{L}} t - c\bm{\widetilde{L}}^{\top} s} \notag\\
   & \quad (e^{c(\bm{\widetilde{L}} + \fractionalLaplacian^{\top} ) \min(t, s)} - \bm{I}) (\bm{\widetilde{L}} + \bm{\widetilde{L}}^{\top} )^{-1}.
\end{flalign}
Or, when the matrix $\bm{\widetilde{L}}$ is self-adjoint (the graph is undirected), as 
\begin{flalign}
    & \bm{\mu}(t) =e^{-c\bm{\widetilde{L}} t} \bm{u}(0),\\
    & \Cov[\bm{u}(t), \bm{u}(s)] = \frac{\sigma^2}{2c} \left(e^{-c \bm{\widetilde{L}} |t - s|} - e^{-c \bm{\widetilde{L}} (t + s)}\right)\bm{\widetilde{L}}^{-1} \notag.
\end{flalign}
\end{FrameProposition}

The kernel is parameterized by diffusivity $c$, variance $\sigma$, and parameters of the fractional Laplacian $\nu$ and $\kappa$. We extend this result to the case of matrix-scaled white noise:
\smallskip
\begin{FrameProposition}
\label{proposition:shek_matrix}
\textbf{SHEK with matrix-scaled noise.} A scaled extension of SHEK can be derived from a stochastic heat equation on graphs with a matrix-scaled white noise:
\begin{equation}
    \frac{\mathrm{d} \bm{u}}{\mathrm{d} t} = -c \bm{\widetilde{L}} \bm{u} + \bm{\Sigma} \dt{\bm{W}}_t.
\end{equation}
The solution for undirected graphs is defined by a Gaussian process with covariance matrix in the following form:
\begin{equation}
    \Cov\left[\bm{u}(t), \bm{u}(s)\right] = \bm{P}^{*} \bm{C}(t, s) \bm{P},
\end{equation}
where $\bm{P}$ is a unitary matrix: 
\begin{equation}
    \bm{P} \bm{\widetilde{L}} \bm{P}^{*} = \operatorname{diag}({\lambda_1, \ldots, \lambda_{|V|}}).
\end{equation}
The matrix $\bm{P}$ exists because $\bm{\widetilde{L}}$ is normal and positive definite.
$\bm{C}(t, s)$ is defined for $t \geq s$ as:
\begin{multline}
    \bm{C}(t, s)_{i,j} = \frac{1}{c}\frac{(\bm{P} \bm{\Sigma} \bm{\Sigma}^{\top} \bm{P}^{*})_{i,j}}{\lambda_i + \lambda_j}  \\
    \times \big(\exp(- c \lambda_i |t - s|) - \exp(- c(\lambda_i t + \lambda_j s))\big).
\end{multline}
\end{FrameProposition}
For a directed graph, the stationary covariance matrix $\bm{C}_{*}$ can be found from the Lyapunov equation
\begin{equation}
    \bm{\widetilde{L}} \bm{C}_{*} + \bm{C}_{*} \bm{\widetilde{L}}^{\top} = \bm{\Sigma} \bm{\Sigma}^{\top}.
\end{equation}

\subsection{Wave Equation}
Spreading waveforms in physics, such as mechanical waves (e.g., sound waves or seismic waves) as well as light waves, are commonly described by the wave equation:
\begin{equation}
    \frac{\mathrm{d}^{2} f}{\mathrm{d} t^2} = c^2 \Delta f. 
\end{equation}
This PDE can be considered on graphs as follows:
\begin{equation} \label{equation:wave_equation}
    \frac{\mathrm{d}^{2} \bm{u}}{\mathrm{d} t^2} = -c^2 \bm{L} \bm{u}.
\end{equation}
The solution of this equation for undirected graphs has the form
\begin{equation}
\begin{split}
    \bm{u}(t) = \frac{1}{c} \sqrt{\bm{L}^{+}} \sin(c \sqrt{\bm{L}} t) \dt{\bm{u}}(0) + \cos(c \sqrt{\bm{L}}t) \bm{u}(0).
\end{split}
\end{equation}

The solution of the stochastic wave equation cannot be used directly as a GP kernel (as it was with the heat equation), but this result is necessary for the derivation of a GP kernel from stochastic wave equations on graphs.

\subsection{Stochastic Wave Equation}
\label{section:wave_equation}
In this section, we consider a stochastic extension of the wave equation on graphs. Adding stochasticity allows us to derive a GP kernel that will include prior information about wave nature of the underlying process.
\smallskip
\begin{FrameProposition}
The \textbf{stochastic wave equation kernel (SWEK)} on undirected graphs is defined by the second-order matrix differential equation 
\begin{equation} \label{equation:stochastic_wave_equation}
    \frac{\mathrm{d}^2 \bm{u}}{\mathrm{d} t^2} = -c^2 \bm{\bm{\widetilde{L}}} \bm{u} + \sigma \dt{\bm{W}}_t,
\end{equation}
and a solution to this equation for undirected graphs can be expressed by the Gaussian process:
\begin{flalign}
    & \bm{u}(t) \sim \mathcal{GP}(\bm{\mu}, \Cov [\bm{u}(s), \bm{u}(t)], \quad \text{with}\\
    & { \bm{\mu}(t) = \frac{1}{c} \bm{\widetilde{L}}^{-\frac{1}{2}} \sin(c \sqrt{\bm{\widetilde{L}}} t) \dt{\bm{u}}(0) {+} \cos(c \sqrt{\bm{\widetilde{L}}} t ) \bm{u}(0)},\nonumber \\
    & \Cov [\bm{u}(s), \bm{u}(t)] = \sigma^2 \bm{\Theta}^{-2}\bigg(\cos(\bm{\Theta} (t - s)) \, \min(t, s) -\nonumber \\ &\quad\quad \frac{1}{2} \cos(\bm{\Theta} \max(t, s)) \sin(\bm{\Theta} \min(t, s)) \bm{\Theta}^{-1}\bigg),\nonumber
\end{flalign}
where $\bm{\Theta} = c \sqrt{\bm{\widetilde{L}}}$.
\end{FrameProposition}

We can observe that eigenvalues of the kernel oscillate with different frequencies. It makes this kernel potentially useful for periodic processes on graphs or discretizations of continuous periodic processes with a graph lattice.

The stochastic wave equation kernel can be extended to directed graphs similarly by solving the graph wave equation through the eigenvectors of $\bm{\widetilde{L}}$.

In a single-vertex graph, this kernel coincides with the product of Brownian motion kernel and cosine kernel.

\subsection{Sampling in Graph GP Models}
\label{section:sampling}
Many methods for sampling from GPs are available. A common method involves a Cholesky decomposition of the covariance matrix. The computational cost of sampling is a sum of $O(n^3)$ on Cholesky factorization and $O(n^2)$ (matrix-vector multiplication) for each sample, where $n$ is the number of points.

Many other sampling approaches are based on the expansion of the random field (process) as a series of basis functions.
Those methods include Karhunen--Loeve expansion using the eigenfunctions of a covariance function, circulant embedding methods based on Fourier basis \citep{graham2018analysis}, and a hierarchical matrix approximation of the covariance matrix \citep{feischl2018fast}. The SPDE based approach \citep{lindgren2011explicit} requires solving an SPDE using the finite element method, which approximates the GP with a Gaussian Markov random field (GMRF).

The SPDE framework allows for using the numerical solvers of SPDEs, for example, the Ruge--St\"uben solver \citep{ruge1987algebraic} for the spatial case or Euler--Maruyama \citep{higham2002strong} for the spatio-temporal case. This idea especially shines in the graph domain because the Laplacian matrix allows considering these equations as matrix equations; thus, it allows using all of the existing tools for numerical solutions of matrix equations. Moreover, in principle, continuous domains can be discretized via graph representations, and then numerical solvers can be used for sampling from GPs over continuous domains. We consider this opportunity as one more justification for the SPDE framework and leave its development for future works.

\subsection{Visualizations of the Kernels}
\label{section:kernel_properties}
For a deeper understanding of the properties of the kernels and a more pictorial tour of their properties, we study visualizations of the kernels on a simple graph with different values for the kernel hyperparameters.

\begin{figure*}
\vspace{-0.5em}
\hspace{1cm}\begin{adjustbox}{trim=0 0 0 2.5cm}{\begin{tikzpicture}[scale=0.92]
    \useasboundingbox (0,0) rectangle (0,4);
    \node[shape=circle,draw=black!70, line width=2pt] (A) at (0, 6) {$f_1(t)$};
    \node[shape=circle,draw=black!70, line width=2pt] (B) at (0, 4) {$f_2(t)$};
    \node[shape=circle,draw=black!70, line width=2pt] (C) at (0, 2) {$f_3(t)$};
    \path [-,line width=2pt,draw=black!70](A) edge (B);
    \path [-,line width=2pt,draw=black!70](B) edge (C);
\end{tikzpicture}\vspace{6cm}}\end{adjustbox}\vspace{-0.5cm}%
 \includegraphics[width=0.5\textwidth]{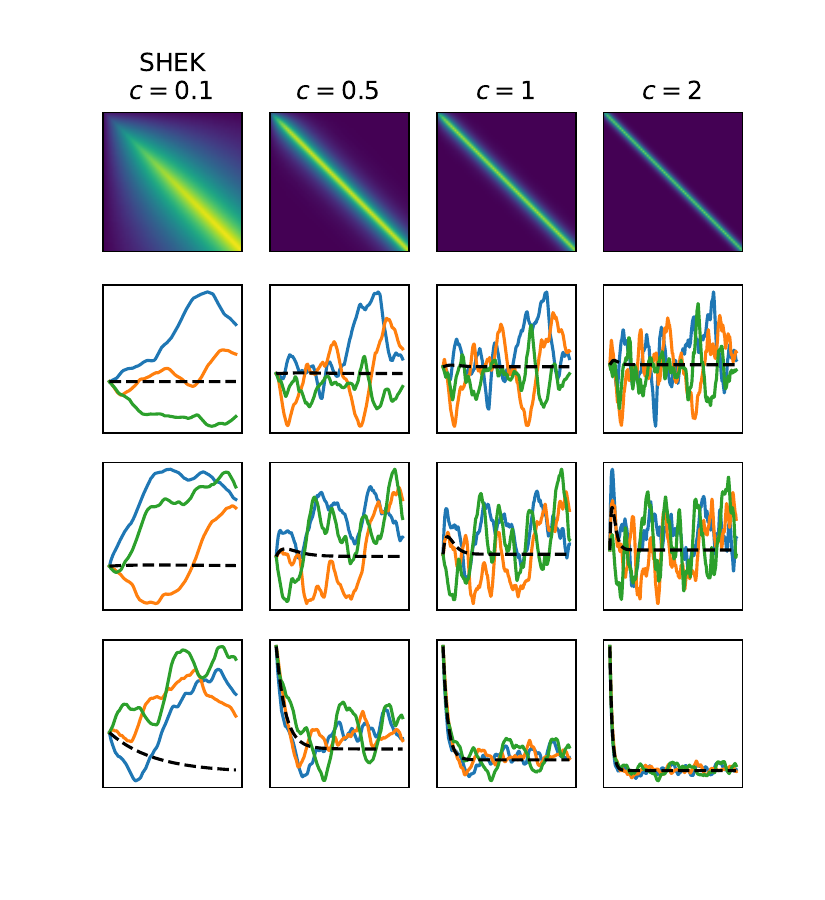}%
 \hspace{-1cm}\includegraphics[width=0.5\textwidth]{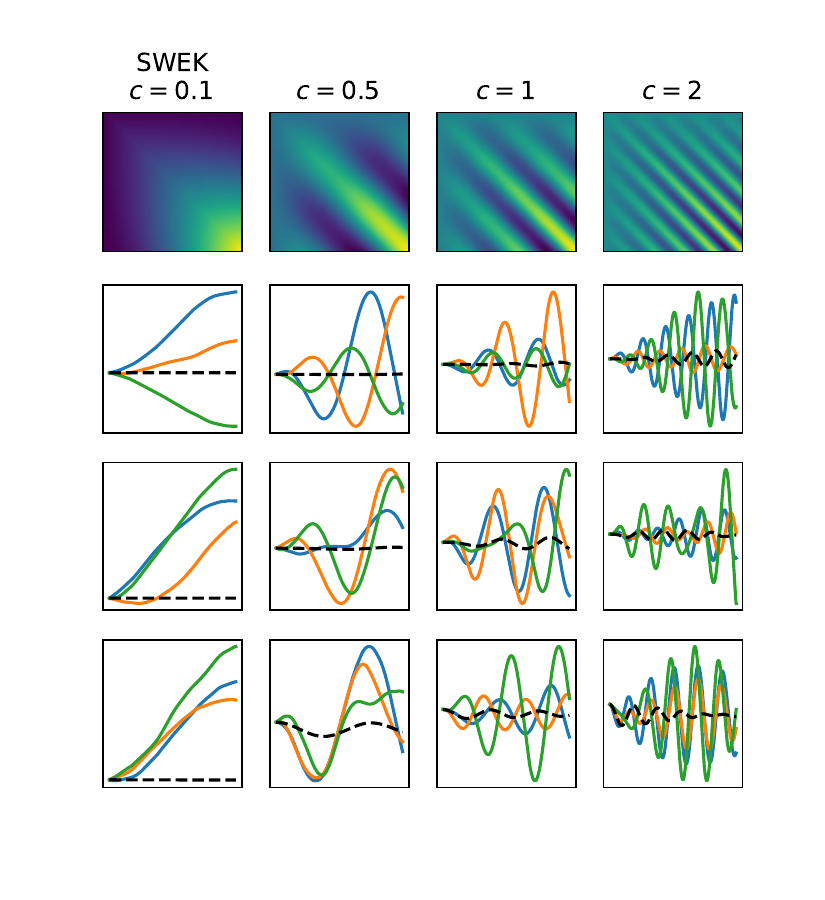}
     \caption{Temporal visualizations of SHEK (heat, left) and SWEK (wave, right) on a linear three-node graph. The first row shows the temporal part of the covariance matrix (summed over the graph vertices at each timepoint). The following three rows show mean (black) and samples (colored lines) as a function of time at each of the nodes, conditioned on $y(t=0) = (0, 0, 10)$, for different values of the hyperparameter $c$.}
    \label{figure:graph_visualizations}
\end{figure*}
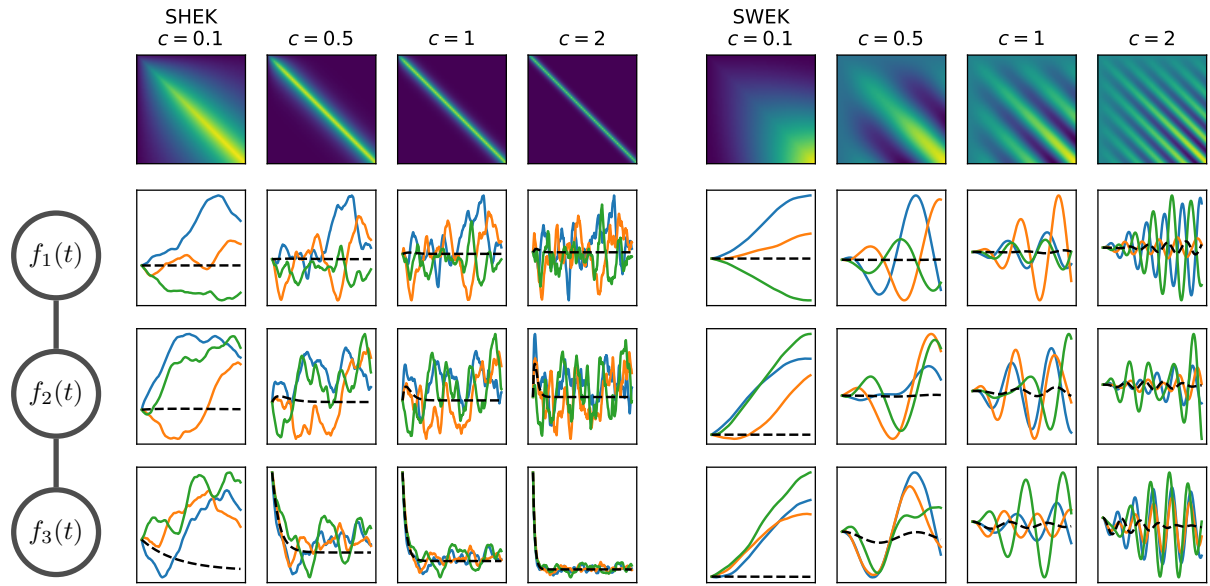

In \cref{figure:graph_visualizations}, we show the visualizations of the proposed kernels on a toy model: a line graph of three vertices.
The first row shows the temporal part of the covariance matrix (summed over the graph vertices at each timepoint). The following three rows show mean (black) and samples (colored lines) as a function of time at each of the nodes, conditioned on $y(t=0) = (0, 0, 10)$, for different values of the hyperparameter $c$ ($\nu=3/2$, $\kappa=1$).
For the temporal component, SHEK shows the expected behavior for a diffusion process: covariance is high between nearby time points and decreases with the temporal gap, whereas SWEK shows periodic dependence between temporal components (periods of high covariance are alternating with periods of low covariance). Additionally, stochasticity accumulates with time resulting in higher covariance for later timestamps (analogously to Brownian motion).

\section{EXPERIMENTS}
\label{section:experiments}
To highlight practical applicability of the proposed methodology and benchmark against alternative approaches, we evaluate the described kernels on graph spatio-temporal applications:
heat distribution over a one-dimensional line (\cref{section:heat_transfer_dataset}), 
 spreading of {COVID-19} across the United States (\cref{section:covid19_distribution_dataset}),
and spreading of chickenpox over Hungarian counties (\cref{section:chickenpox}).

We used Gaussian process regression with Gaussian likelihood as a method for fitting the models. We maximized the log marginal likelihood with respect to kernel hyperparameters ($\sigma$ and $c$ for SWEK and SHEK) with the L-BFGS optimizer \citep{liu1989limited}.  All the experiments were implemented using the GPflow \citep{GPflow2017} library in Python.

\textbf{Extrapolation.} Our experiments were repeated for different training and testing intervals (we call them validation rounds) for a given data set using sliding window backtesting. If we denote number of training timepoints as $N_\text{train}$, number of testing timepoints as $N_\text{test}$, and by $t^{r}$ the starting time for the $r$-th validation round, then sliding backtesting uses time interval $[t^{r}, \ldots, t^{r} + N_\text{train}]$ as the training data, and $[t^{r} + N_\text{train} + 1, \ldots,  t^{r} + N_\text{train} + N_\text{test}]$ as test data. 

\textbf{Interpolation.}
To evaluate interpolation properties of the methods, we randomly selected ten percent of the time interval $[t^r, \ldots, t^{r} + N_\text{train} + N_\text{test}]$ as testing data set and used the rest for training.

\textbf{Evaluation.}
In forecasting problems, modelling may result in a high variance of the generalization error across the validation rounds. Thus, the comparison of generalization error over the rounds by confidence intervals can be inaccurate. The generalization errors should be evaluated with a thorough point-wise comparison over each evaluation. To make statistically significant conclusions, we used the Diebold--Mariano test \citep{diebold2002comparing}. The variance of the generalization error is also an important model property that shows the method's robustness over different validation rounds. We tested both interpolation and extrapolation performance and report mean absolute error ($\text{MAE}_{\text{int}}$ and $\text{MAE}_{\text{ext}}$) or mean absolute percentage error ($\text{MAPE}_{\text{int}}$ and $\text{MAPE}_{\text{ext}}$). We also report 95\% confidence intervals over validation rounds.

\subsection{Heat Transfer Data Set}
\label{section:heat_transfer_dataset}
As a simple spatio-temporal process, we considered heat transfer in 1D on a line. We parameterized the process with conductivity $k$, and considered a fundamental solution of the heat distribution process: $\Phi(x, t) = {5}/({4 \pi k t}) \exp(-{x^2}/{4 k t})$. We modelled the heat transfer distribution for a one-dimensional segment and discretized it as a linear graph with 21 vertices.

We compared separable and non-separable graph kernels for temporal signals. We used a continuous time-separable kernel as a reference model. We used 50 timestamps for training data and 10 time\-stamps as test data; we performed ten iterations of backtesting validation with different random seeds and different starting time points. The results are presented in \cref{table:heat_transfer}; we report extrapolation and interpolation MAE with 95\% confidence intervals, marking the top results of the graph kernels in bold. We observed that the non-separable stochastic heat kernel outperformed separable kernels on the extrapolation task, but the product graph Mat\'ern kernel and RBF outperformed it on the interpolation task. As expected, exponential kernel over continuous space outperformed graph discretization for extrapolation but did not outperform the product of graph Mat\'ern kernel and RBF over time.

\begin{table}[t!]
\caption{Interpolation/extrapolation performance for the heat distribution problem over a line ($\text{MAE} \times 100$).}
\label{table:heat_transfer}
\setlength{\tabcolsep}{5pt}  
\footnotesize
\centering
\begin{tabularx}{\columnwidth}{lcc}
\toprule
Kernel & $\text{MAE}_{\text{int}}$ & $\text{MAE}_{\text{ext}}$ \\
\midrule
Laplacian(G) $\times$ RBF(T) & 0.42 $\pm$ 3.21 & 4.75 $\pm$ 0.36 \\
Mat\'ern-1/2(G) $\times$ RBF(T) & \textbf{0.05} $\pm$ \textbf{0.05} & 2.81 $\pm$ 4.12\\
SHEK(G, T) & 0.13 $\pm$ 0.11 & \textbf{1.29} $\pm$ \textbf{0.06}\\
\midrule
Exp($\mathbb{R}$) $\times$ Exp(T) & \textit{0.12} $\pm$ \textit{0.12} & \textit{1.02} $\pm$ \textit{0.05} \\ 
\bottomrule
\end{tabularx}
\end{table}

\begin{table*}[t]
\caption{Summary of the interpolation ($\text{int}$) and $k$-weeks extrapolation ($\text{ext-}k$) results in terms of mean absolute error (MAE) and mean absolute percentage error (MAPE) on COVID-19 and Hungarian chickenpox data sets. Note that the DCRNN model does not support interpolation.}
\label{table:mae_covid_chickenpox}
\setlength{\tabcolsep}{6pt}  
\centering
\begin{tabularx}{\textwidth}{lccccc}
\toprule
 & \multicolumn{3}{c}{\sc Chickenpox in Hungary} & \multicolumn{2}{c}{\sc COVID-19 in the US} \\
\cmidrule(lr){2-4}\cmidrule(lr){5-6}
\multicolumn{1}{c}{Model} & $\text{MAE}_{\text{int}}$ & $\text{MAE}_{\text{ext-4}}$ & $\text{MAE}_{\text{ext-6}}$ & $\text{MAPE}_{\text{int}}$ & $\text{MAPE}_{\text{ext-2}}$ \\

\midrule
Laplacian(G) $\times$ RBF(T) & 33.95 $\pm$ 1.64 & 34.25 $\pm$ 1.68  & 46.45 $\pm$ 4.93 & 1.58 $\pm$ 0.53 & 0.76 $\pm$ 0.03  \\
Mat\'ern-3/2(G) $\times$ RBF(T) & \textbf{13.82} $\pm$ \textbf{0.72} & \textbf{26.45} $\pm$ \textbf{4.05} & 32.34 $\pm$ 3.07 & 0.17 $\pm$ 0.02 & \textbf{0.27} $\pm$ \textbf{0.08}\\
DCRNN(G,T) & --- & 27.96 $\pm$ 3.65 & \textbf{30.45} $\pm$ \textbf{2.19} & --- & 0.56 $\pm$ 0.07 \\
SHEK(G, T) ($\nu=\tfrac{1}{2}$, $\kappa=1$) &  14.81 $\pm$ 0.91 & \textbf{26.46} $\pm$ \textbf{3.73} &  \textbf{30.65} $\pm$ \textbf{2.51} & \textbf{0.16} $\pm$ \textbf{0.03} & \textbf{0.25} $\pm$ \textbf{0.04}  \\
\bottomrule
\end{tabularx}

\end{table*}

\subsection{COVID-19 Distribution Across the US}
\label{section:covid19_distribution_dataset}
As a real-world use-case of the proposed method, we considered the prediction of COVID-19 distribution and spread across the USA. 
We used data about COVID-19 cases and deaths published by \citet{covid_dataset}. We aggregated the number of cases by week for each state. Moreover, we handcrafted a graph that represents the geographical neighbors of the states in the US. Each node of the graph represents a state, and two nodes are connected with an undirected edge if the corresponding states share a common border. The approach can be extended to other graphs that arise from, for example, traffic links or information about the flights. As a modelling target for the extrapolation task, we aimed to predict how many cases there will be in each state during the next two weeks. For numerical stability, we predicted the natural logarithm of this value. We used 33 weeks as training data and estimated the number of cases for the subsequent two weeks. We ran all experiments ten times with a sliding starting point and different random seeds. The goal of the experiment was to show that the proposed kernels can achieve good results within the GP framework. We argue that GPs are particularly useful in this case because of their excellent interpolation properties and the possibility to assess the uncertainties, which is crucial for decision-making. We compared the results with a graph neural network (GNN), for which we used a diffusion convolutional recurrent neural network (DCRNN) layer \citep{li2017diffusion} followed by a fully connected layer. We used the implementation from the library `PyTorch Geometric Temporal' \citep{rozemberczki2021pytorch}. We also deepened the architecture but the results remained largely unaffected. For this reason, we report the results for the simplest architecture.

The results are shown in \cref{table:mae_covid_chickenpox}. We can see that the models converged to a relatively low MAPE, and SHEK showed the best performance. This shows that the proposed methods are useful for the task we consider. The results on the interpolation task showed better performance for SHEK (DM-test, $p=0.1$). The results on extrapolation performance are equal for considered separable and non-separable cases. However, SHEK's performance was more stable across validation rounds on the extrapolation task, resulting in two times smaller 95\% confidence intervals of generalization error. When the extrapolation period was increased to four and six weeks, SHEK performed consistently better than other kernels: four weeks period MAE of SHEK($\nu=1/2$) was less than MAE of the product of Mat\'ern(1/2) and RBF (DM-test, $p < 0.01$), as well as on six week period (DM-test, $p < 0.05$).

\begin{figure}[htbp!]
    \centering
    \begin{subfigure}[b]{0.47\textwidth}
    \centering\scriptsize
      {\includegraphics[width=\linewidth]{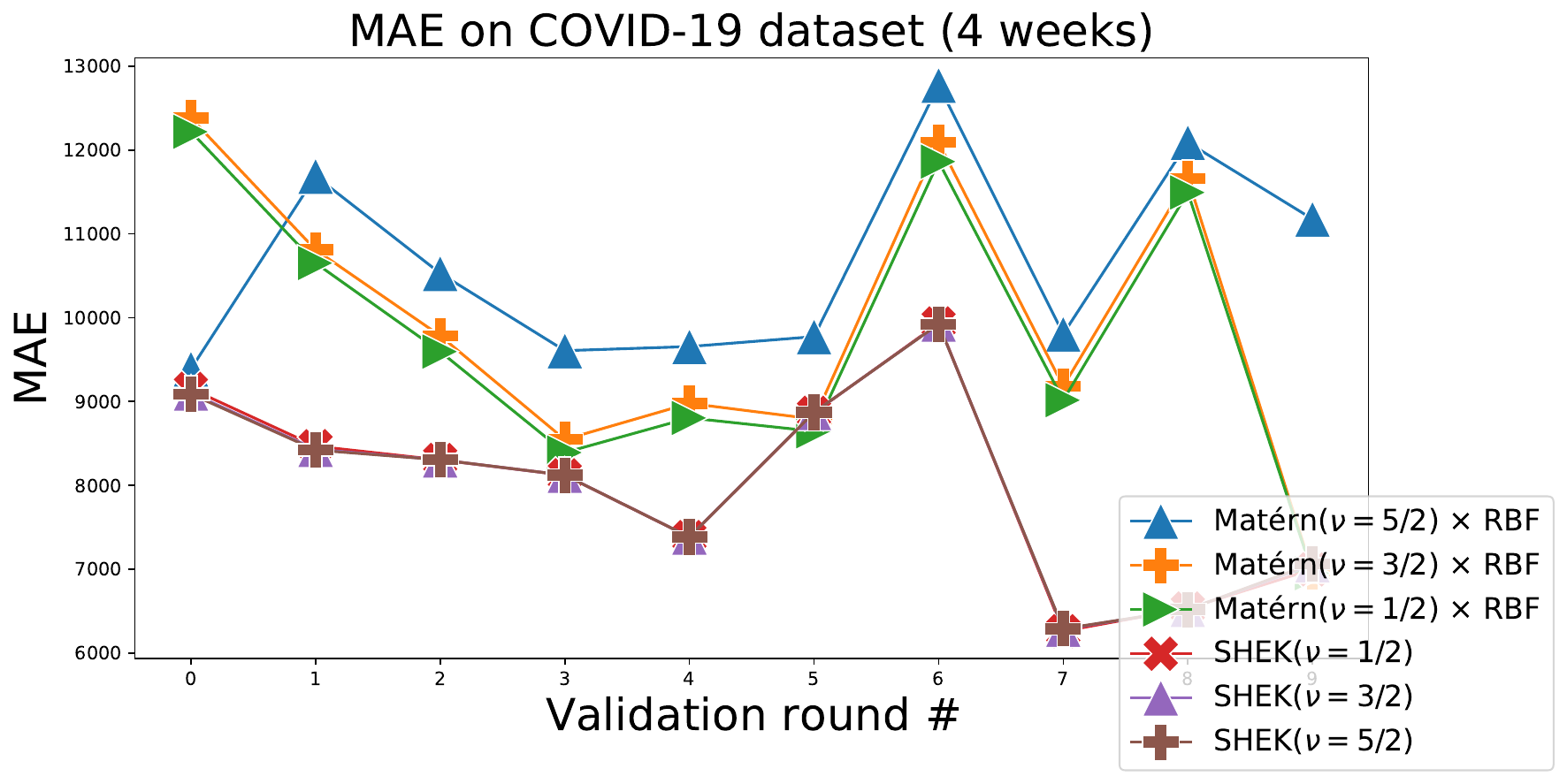}}
      \caption{Four weeks extrapolation.}
      \label{figure:covid_pointwise_comparison_4}
    \end{subfigure}
    \hfill
    \begin{subfigure}[b]{0.47\textwidth}
    \centering\scriptsize
      {\includegraphics[width=\linewidth]{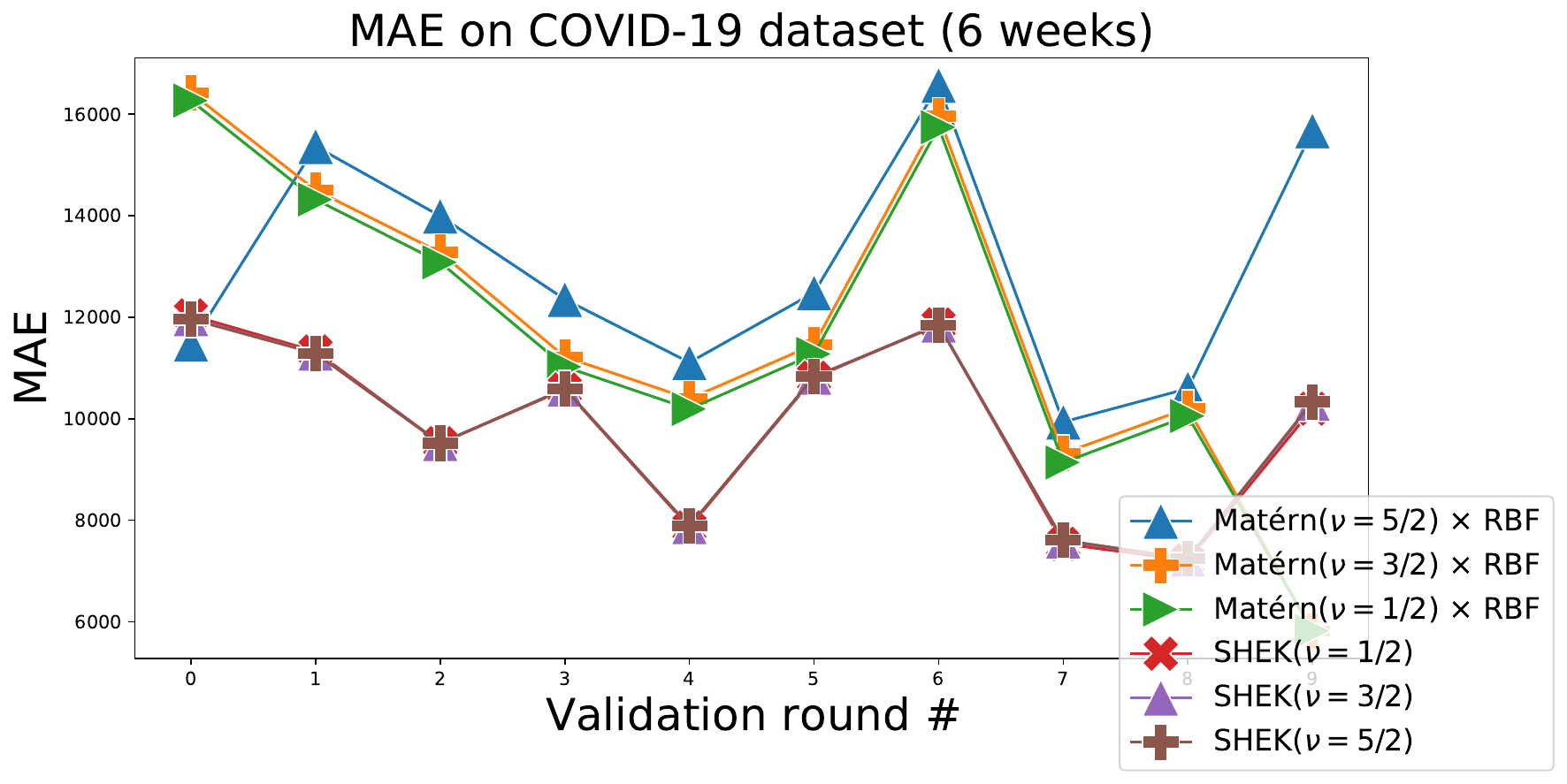}}
      \caption{Six weeks extrapolation.}
      \label{figure:covid_pointwise_comparison_6}
    \end{subfigure}

\caption{Visualization of extrapolation evaluation with extended extrapolation periods of four and six weeks on COVID-19 dataset.}
\label{figure:appendix_MAE_Covid}
\end{figure}

\textbf{Experiment with a flight graph.} Additionally, We conducted an experiment to check whether non-spatial graph information can be included in the Covid-19 modeling task. We collected information about flights between the states and used this graph instead of geographical adjacencies. Results showed MAE statistically better for non-separable ($p=0.05$, DM-test); MAPE showed no statistically significant difference between separable and non-separable kernels in this case.

\subsection{Hungary Chickenpox Data Set}
\label{section:chickenpox}
We evaluated our methods on a data set of chickenpox cases in Hungary \citep{rozemberczki2021chickenpox}. The data set contains the weekly aggregated numbers of cases of chickenpox in Hungarian counties and Budapest from 2005 until 2015. The spatial relation of counties is represented as a graph with each county as a node (in total 20 nodes) and unweighted edges that indicate adjacency of the counties (in total 61 edges). This data set has a different structure than the COVID-19: the epidemic has strong seasonality and does not have periods of fast exponential growth. 

We observed similar results to the COVID-19 case: GP approaches outperformed the GNN approach, SHEK has shown comparable performance with separable kernel on the extrapolation task with more stable performance across the validation rounds. On the long-term (six weeks) extrapolation task, DCRNN and SHEK outperformed other approaches. The separable kernel has shown better performance on the interpolation task than SHEK (DM-test, $p=0.05$). We repeated the experiment for the longer extrapolation periods: four and six weeks and observed that SHEK performed more stable and consistently better than separable kernels (DM-test, $p=0.1$ for four weeks and $p=0.05$ for six weeks). We also experimented with other types of separable kernels and selected Mat\'ern-3/2 plus RBF as a consistently accurate combination on interpolation and extrapolation tasks. We provide full results of the evaluation and point-wise comparison in Appendix~\ref{section:appendix_chickenpox_details}.

\section{DISCUSSION \& CONCLUSION}
\label{section:conclusion}
We considered spatio-temporal problems on graphs. We showed how these can be addressed by product-separable kernels and, then, introduced a framework to derive spatial and spatio-temporal graph kernels based on SPDEs. As concrete instances, we considered the stochastic heat and wave equations on graphs, and solved them to arrive at novel types of graph kernels: the stochastic heat equation kernel (SHEK) and the stochastic wave equation kernel (SWEK). We compared GPs with the proposed kernels against DCRNNs, and demonstrated that our approach outperforms graph neural networks on some machine learning tasks.

Following our method for deriving kernels from SPDEs, this work creates a possibility for direct use of additional prior information in graph problems by creating new flexible kernels on graphs. These methods can be applied to other types of equations resulting in either an analytical solution or an SPDE that can be numerically solved. It may lead to the more frequent use of probabilistic methods on graphs overcoming predictive power issues with the pre-existing kernels.

Our work opens up new avenues of research, such as the solution of other types of equations on graphs or discretization of continuous kernels using graph approximation. Moreover, it gives machine learning tools that can be used for essential problems on graphs, such as epidemic modelling. As a side contribution of our work, we constructed a timely data set of COVID-19 cases over a graph that can be used to evaluate temporal machine learning problems on graphs.

In future work, the scalability of the methods to larger graphs has to be explored. Possible approaches include state space models, numerical solvers for SPDEs, and sparse methods.
\begin{flushleft}
Source code and data for the experiments are available at: \url{https://github.com/AaltoPML/spatiotemporal-graph-kernels}
\end{flushleft}
\acknowledgments{This work was supported by the Academy of Finland Flagship Programme Finnish Center for Artificial Intelligence FCAI, Academy of Finland grants 339730, 341763 and 324345, and UKRI Turing AI World-Leading Researcher Fellowship, EP/W002973/1. We also acknowledge the computational resources provided by the Aalto Science-IT Project from Computer Science IT. We thank Ivan Yashchuk and Ivan Yaroslavtsev for their helpful comments.}

\bibliography{main}
\bibliographystyle{apalike}

\onecolumn
\appendix
\nipstitle{Supplementary Material for \\Non-separable Spatio-temporal Graph Kernels via SPDEs}
\thispagestyle{empty}
\numberwithin{equation}{section}
\numberwithin{definition}{section}
\numberwithin{theorem}{section}
\numberwithin{proposition}{section}

\setcounter{table}{0}
\renewcommand{\thetable}{S\arabic{table}}
\setcounter{figure}{0}
\renewcommand{\thefigure}{S\arabic{figure}}

\section{Preliminaries}
In our proofs, we rely on the following theorem.
\begin{theorem}[It\^o isometry, \cite{oksendal2013stochastic}]
\begin{equation}
    \EX\left[\left(\int_S^T f(t, w)\,\mathrm{d}W_t\right)^2\right] =\EX\left[\int_S^T f^2(t, w) \,\mathrm{d} t\right],
\end{equation}
for all $f \in \mathcal{V}\left(S, T\right)$.
\end{theorem}

\begin{corollary}
\begin{equation}
    \EX\left[\left(\int_0^t X_t \,\mathrm{d} W_t\right) \left(\int_0^t Y_t \,\mathrm{d}W_t\right)\right] = \EX\left[\int_0^t X_t Y_t \,\mathrm{d}t\right],
\end{equation}
where $X_t$ and $Y_t$ are stochastic processes adapted to natural filtration. This property will be used actively in derivations of covariances from SDEs.
\end{corollary}

\begin{lemma}[Integration of the matrix exponential]
\begin{equation}
   \int_0^T e^{\bm{A}t}\,\mathrm{d}t = \left(e^{\bm{A}T} - \bm{I}\right) \bm{A}^{-1},
\end{equation}
where $\bm{A}$ is a nonsingular matrix.
\end{lemma}

\section{Graph Kernels}
\label{section:appendix_graph_kernels}
In this section, we introduce spatial kernels on graphs derived by the SPDE framework.
\begin{proposition}[Laplace's equation in kernel form]
\label{theorem:appendix_kernel_laplace_equation}
Consider a signal (process) $\bm{v} \in \mathbb{R}^{|V|}$ on a graph $G$, where $\bm{L}$ is the Laplacian of $G$ and the signal is described with the Laplace SPDE on the graph:
$-\bm{L} \bm{v} = \bm{w}$.
Then the solution $\bm{v}$ can be described as a Gaussian process: 
\begin{equation} \label{equation:appendix_laplacian_kernel}
    \bm{v} \sim \mathcal{N}\left(\bm{0}, \left(\bm{L}^{*} \bm{L}\right)^{+}\right).
\end{equation}
\end{proposition}

\begin{proof}
By writing out
\begin{equation}
  \bm{v} = -\bm{L}^{+}\bm{w},
\end{equation}
and noting that the covariance matrix of white noise $\bm{w}$ is $\bm{I}$, the covariance of the random variable $\bm{u}$ is
\begin{equation}
    \Cov\left[-\bm{L}^{+} \bm{w}\right] = \bm{L}^{+} \Cov\left[\bm{w}\right] {\bm{L}^{+}}^{\top} = \left(\bm{L}^{\top} \bm{L}\right)^{+}.
\end{equation}
\end{proof}

\begin{proposition}[Mat\'ern kernels on graphs] Mat\'ern kernels on graphs are described with the equation 
\begin{equation} \label{equation:appendix_matern_kernel_equation}
  \bigg(\frac{2 \nu}{\kappa^{2}} \bm{I} + \bm{L}\bigg)^{\frac{\nu}{2}} \bm{v}=\bm{w},
\end{equation}
and have the form
\begin{equation}
\bm{K}_{XX} = \bigg[\bigg(\bigg(\frac{2\nu}{\kappa^2} \bm{I} + \bm{L}\big)^{-\frac{\nu}{2}}\bigg)^{\top}  \bigg(\frac{2\nu}{\kappa^2} \bm{I} + \bm{L}\bigg)^{-\frac{\nu}{2}}\bigg]^{-1}.
\end{equation}

For the self-adjoint Laplacian $\bm{L}$:
\begin{equation}
\bm{K}_{XX} = \bigg(\frac{2\nu}{\kappa^2} \bm{I} + \bm{L}\bigg)^{\nu}.
\end{equation}
Thus, the solution of Eq.~\eqref{equation:appendix_matern_kernel_equation} is
\begin{equation}
    \bm{v} \sim \mathcal{N}\bigg(\bm{0}, \bm{K}_{XX}\bigg).
\end{equation}
\end{proposition}

\begin{proof}
Let $\bm{A} = \bigg(\frac{2 \nu}{\kappa^{2}} \bm{I} + \bm{L}\bigg)^{\frac{\nu}{2}}$, then $\bm{v} = \bm{A}^{-1} \bm{w}$. Now, we can derive a covariance matrix, analogously as in Proposition~\ref{theorem:appendix_kernel_laplace_equation}.
\end{proof}

\section{Stochastic Heat and Wave Equations on Graphs}
\label{section:appendix_wave_and_stochastic_heat}
\subsection{Heat Equation}
\begin{proposition}[Heat equation] The heat equation on graphs is defined by the second order matrix differential equation 
\begin{equation} \label{equation:appendix_heat_equation}
    \frac{\mathrm{d} \bm{u}}{\mathrm{d} t} = -c\, \bm{L}\bm{u},
\end{equation}
and the solution of this equation is:
\begin{equation}
\begin{split}
    \bm{K}_{H} = e^{-c \bm{L} t} \bm{u}(0).
\end{split}
\end{equation}
\end{proposition}

\begin{proof}
    The solution of the differential equation is similar to the homogeneous initial value problem in the continuous case and can be written in terms of a matrix exponential:
\begin{equation}
    \bm{u}(t) = e^{-c \bm{L}t} \bm{u}(0).
\end{equation}
\end{proof}

For stochastic heat and wave equations, we will be using the pseudo-differential operator $\bm{\widetilde{L}}$ instead of the graph Laplacian because this operator is more flexible than $L$ (for $\nu = 1$ and $\kappa \rightarrow \infty$, it coincides with $L$), and because it possesses non-local properties as was shown by \citet{benzi2020non}.

\begin{proposition}[Stochastic heat equation on graphs] The stochastic heat equation can be defined on a graph by adding spatio-temporal white noise, or for convenient integration, as the {\em formal} derivative of Wiener process $\dt{\bm{W}}_t$:
\begin{equation} \label{equation:appendix_stochastic_heat_equation_on_graph}
    \frac{\mathrm{d} \bm{u}}{\mathrm{d} t} = -c\bm{\widetilde{L}} \bm{u} + \sigma \dt{\bm{W}}_t.
\end{equation}

The solution can be defined as a Gaussian process:
\begin{flalign}
  \bm{u}(t) &\sim \mathcal{GP}(\bm{\mu}(t), \Cov [\bm{u}(s), \bm{u}(t)]), \quad \text{with} \notag \\
   \bm{\mu}(t) &= e^{-c\bm{\widetilde{L}} t} \bm{u}(0), \notag\\
   \Cov[\bm{u}(t), \bm{u}(s)] &= \frac{\sigma^2}{c} e^{-c\bm{\widetilde{L}} t - c\bm{\widetilde{L}}^{\top} s}  \left(e^{c(\bm{\widetilde{L}} + \fractionalLaplacian^{\top} ) \min(t, s)} - \bm{I}\right) (\bm{\widetilde{L}} + \bm{\widetilde{L}}^{\top} )^{-1}.
\end{flalign}
Or, when the matrix $\bm{\widetilde{L}}$ is self-adjoint (the graph is undirected), as 
\begin{flalign}
    \bm{\mu}(t) &= e^{-c\bm{\widetilde{L}} t} \bm{u}(0),\\
    \Cov[\bm{u}(t), \bm{u}(s)] &= \frac{\sigma^2}{2c} \left(e^{-c \bm{\widetilde{L}} |t - s|} - e^{-c \bm{\widetilde{L}} (t + s)}\right)\bm{\widetilde{L}}^{-1} \notag.
\end{flalign}
\end{proposition}

\begin{proof}
Let $\bm{\Gamma} = c\bm{\widetilde{L}}$.
Equation~\eqref{equation:appendix_stochastic_heat_equation_on_graph} is a matrix differential equation and a solution can be written in the form:
\begin{equation} \label{equation:appendix_stochastic_heat_equation_solution}
    \bm{u}(t) = e^{-\bm{\Gamma} t} \bm{u}(0) + \sigma e^{-\bm{\Gamma} t} \int_{0}^{t} e^{\bm{\Gamma} s} \,\mathrm{d} \bm{W}_s.
\end{equation}

The solution in Eq.~\eqref{equation:appendix_stochastic_heat_equation_solution} can be expressed as a GP. We give the corresponding mean and covariance as follows.
\begin{equation} \label{equation:appendix_stochastic_heat_equation_solution_mean}
    \EX[\bm{u}(t)] = e^{-\bm{\Gamma} t} \bm{u}(0).
\end{equation}

Using It\^o isometry, we can derive the covariance:
\begin{align}
    \Cov\left[\bm{u}(s), \bm{u}(t)\right] &= \EX\left[\left(\bm{u}(s) - \EX[\bm{u}(s)]\right)\left(\bm{u}(t) - \EX[\bm{u}(t)]\right)^{\top} \right] \nonumber \\
     &=  \EX\left[\left(\sigma e^{-\bm{\Gamma} s} \int_{0}^{s} e^{\bm{\Gamma} \xi} d\bm{W}_{\xi}\right)\, \left(\sigma e^{-\bm{\Gamma}^{\top} t} \int_{0}^{t} e^{\bm{\Gamma}^{\top} \xi} d \bm{W}_{\xi}\right)\right] \nonumber \\
     &= \sigma^2 e^{-\bm{\Gamma} t - \bm{\Gamma}^{T} s} \EX\left[\left(\int_{0}^{s} e^{\bm{\Gamma} \xi} d\bm{W}_{\xi}\right)\left(\int_{0}^{t} e^{\bm{\Gamma}^{T} \xi} d\bm{W}_{\xi}\right)\right] \nonumber \\
     &= {\sigma^2} e^{-\bm{\Gamma} t - \bm{\Gamma}^{T} s} \left(e^{\left(\bm{\Gamma} + \bm{\Gamma}^{\top}\right) \min(t, s)} - \bm{I}\right) \left(\bm{\Gamma} + \bm{\Gamma}^{\top}\right)^{-1}.  \label{equation:appendix_stochastic_heat_equation_solution_covariance}
\end{align}

Or, in the case of self-adjoint $\bm{\widetilde{L}}$ the expression can be simplified:
\begin{equation}
\frac{\sigma^2}{2} \left(e^{-\bm{\Gamma} |t - s|} - e^{-\bm{\Gamma} (t + s)}\right)\bm{\Gamma}^{-1}.
\end{equation}
\end{proof}

\begin{proposition}[Stochastic heat equation on undirected graphs with matrix-scaled white noise] Let us consider the same equation as in the previous theorem but with $\bm{\Sigma}$-scaled white noise on undirected connected graphs:
\begin{equation}
    \mathrm{d} \bm{u} = -c\bm{\widetilde{L}} \bm{u} \, {\mathrm{d} t}  + \bm{\Sigma} \,\mathrm{d} \bm{W_t}.
\end{equation}

The covariance can be derived as follows.
Consider $\bm{\Gamma} = c\bm{\widetilde{L}}$ and a diagonalization of $\bm{\Gamma}$:
\begin{equation}
\bm{P} \bm{\widetilde{L}} \bm{P}^{*} = \bm{P} \bm{\widetilde{L}}^{T} \bm{P}^{+} = \mathrm{diag}(\lambda_1, \lambda_2, \ldots , \lambda_{n}).
\end{equation}

Then, for $t \geq s$:
\begin{equation}
    \Cov\left[\bm{u}(t), \bm{u}(s)\right] = \bm{P}^{*} \bm{C}(t, s) \bm{P},
\end{equation}
where $\bm{C}(t, s)$ is defined as:
\begin{equation}
    \bm{C}(t, s)_{i,j} = \frac{1}{c}\frac{(\bm{P} \bm{\Sigma} \bm{\Sigma}^{\top} \bm{P}^{*})_{i,j}}{\lambda_i + \lambda_j} (\exp(-\lambda_i c |t - s|) - \exp(-c(\lambda_i t + \lambda_j s))).
\end{equation}
\end{proposition}

\begin{proof}
    Let us write covariance between $\bm{u}(t)$ and $\bm{u}(s)$ and apply It\^o isometry
    \begin{align}
        \Cov[\bm{u}(t), \bm{u}^{\top}(s)] &= \EX\left[\big(\bm{u}(t) - \EX[\bm{u}(t)]\big) \big(\bm{u}(s) - \EX[\bm{u}(s)\big)^{\top} \right] \nonumber \\ &= \int_0^{\min(t, s)} \exp(-c \bm{\widetilde{L}}(t - \xi)) \bm{\Sigma} \bm{\Sigma}^{\top} \exp(-c \bm{\widetilde{L}} (s - \xi)) \mathrm{d}\xi.
    \end{align}
    Consider a diagonalization $\bm{P} \bm{\widetilde{L}} \bm{P}^{*} = \bm{L}_d = \operatorname{diag}(\lambda_1, \ldots, \lambda_n)$. Then the covariance can be re-written as 
    \begin{equation}
        \Cov[\bm{u}(t), \bm{u}^{\top}(s)] = \bm{P^{*}} \underbrace{\bigg( \int_0^{\min(t, s)} \exp(-c \bm{L}_d (t - \xi)) \bm{P} \bm{\Sigma} \bm{\Sigma}^{\top} \bm{P^{*}} \exp(-c \bm{L}_d (s - \xi)) \mathrm{d} \xi \bigg)}_{\bm{C}(t, s)} \bm{P}.
    \end{equation}
    Because the matrix $\bm{L}_d$ is diagonal, we can write the $\bm{C}(t, s)_{ij}$ (for $t \geq s$):
    \begin{align}
        \bm{C}(t, s)_{ij} &= (\bm{P} \bm{\Sigma} \bm{\Sigma}^{\top} \bm{P^{*}})_{ij} \int_0^{s} \exp(-c \lambda_i t - c \lambda_j s + c \xi(\lambda_i + \lambda_j))\, \mathrm{d} \xi \nonumber \\ &= \frac{1}{c} \frac{\bm{P} \bm{\Sigma} \bm{\Sigma}^{\top} \bm{P^{*}}}{\lambda_i + \lambda_j} \left(\exp(c \lambda_i |t - s|) - \exp(-c \lambda_i t - c \lambda_j s) \right).
    \end{align}
\end{proof}

\subsection{Wave Equation}
\begin{proposition}[Wave equation on an undirected graph]\label{theorem:appendix_wave_kernel} The wave kernel on undirected graphs is defined by the second order matrix differential equation 
\begin{equation} \label{equation:appendix_wave_equation}
    \frac{\mathrm{d}^2 \bm{u}}{\mathrm{d} t^2} = -c^2 \bm{L} \bm{u},
\end{equation}
and the solution of this equation has a form:
\begin{equation}
\begin{split}
     \bm{u}(t) = \frac{1}{c} \sqrt{\bm{L}^{+}} \sin\left(c \sqrt{\bm{L}} t\right) \dt{\bm{u}}(0) + \cos\left(c \sqrt{\bm{L}}t\right) \bm{u}(0).
\end{split}
\end{equation}
\end{proposition}

\begin{proof}
$\bm{L}$ is a symmetric matrix and, consequently, diagonalizable:
\begin{equation}
    \bm{L} = \bm{P}^{-1} \bm{L}_d \bm{P},
\end{equation}
Then the equation will take a form:
\begin{equation}
    \ddt{\bm{u}} + c^2 \bm{P}^{-1} \bm{L}_d \bm{P} \bm{u} = \bm{0} \Leftrightarrow \bm{P} \ddt{\bm{u}} + c^2 \bm{L}_d \bm{P} \bm{u} = \bm{0}.
\end{equation}
Replacing the variable $\bm{y} \coloneqq \bm{P} \bm{u}$, we will got the following equation:
\begin{equation}
    \ddt{\bm{y}} + c^2 \bm{L}_d \bm{y} = \bm{0}.
\end{equation}

This is equivalent to $n$ independent scalar ODEs:
\begin{equation}
    \ddt{y}_k + c^2 \lambda_k y_k = 0.
\end{equation}

The solution of each of these equation is:
\begin{equation}
    y_k = c_1 \sin \left(c \sqrt{\lambda_k} t\right) + c_2 \cos \left(c \sqrt{\lambda_k} t \right).
\end{equation}

Let the initial conditions of the system be given as $\bm{y}(0), \dt{\bm{y}}(0)$. Then, for $\lambda_k \neq 0$ $c_1$ and $c_2$ can be expressed by
\[
\systeme*{c_1 = \frac{1}{c \sqrt{\lambda_k}} \dt{y}_{k}(0), c_2 = y_{k}(0)}
\]
and $c_1 = c_2 = 0$ for $\lambda_k = 0$.

By doing the reverse substitution, we get the result for u:
\begin{equation}
    \bm{u}(t) = \frac{1}{c} \sqrt{\bm{L}^{+}} \sin\left(c \sqrt{\bm{L}} t\right) \dt{\bm{u}}(0) + \cos\left(c \sqrt{\bm{L}}t\right) \bm{u}(0).
\end{equation}
\end{proof}

\begin{proposition}
The \textbf{Stochastic wave equation kernel (SWEK)} on undirected graphs is defined by second order matrix differential equation 
\begin{equation} \label{equation:appendix_stochastic_wave_equation}
    \frac{\mathrm{d}^2 \bm{u}}{\mathrm{d} t^2} = -c^2 \bm{\widetilde{L}} \bm{u} + \sigma \dt{\bm{W}}_t
\end{equation}
and the solution to this equation can be expressed as a Gaussian process:
\begin{flalign}
    \bm{u}(t) &\sim \mathcal{GP}(\bm{\mu}, \Cov [\bm{u}(s), \bm{u}(t)]\\
    \bm{\mu} &= \frac{1}{c} \bm{\widetilde{L}}^{-\frac{1}{2}} \sin\left(c \sqrt{\bm{\widetilde{L}}} t\right) \dt{\bm{u}}(0) + \cos\left(c \sqrt{\bm{\widetilde{L}}} t \right) \bm{u}(0)\\
    \Cov [\bm{u}(s), \bm{u}(t)] &= \sigma^2 \bm{\Theta}^{-2}\left(\cos(\bm{\Theta} (t - s)) \, \min(t, s) - \frac{1}{2} \cos(\bm{\Theta} \max(t, s)) \sin(\bm{\Theta} \min(t, s)) \bm{\Theta}^{-1}\right).
\end{flalign}
Here, $\bm{\Theta} = c \sqrt{\bm{\widetilde{L}}}$.
\end{proposition}
\begin{proof}
    The homogeneous solution for wave equation is given in Theorem~\ref{theorem:appendix_wave_kernel}: $$\bm{u}(t) = \bm{\widetilde{L}}^{-\frac{1}{2}} \sin\left(c \sqrt{\bm{\widetilde{L}}} t\right) c_1 + \cos\left(c \sqrt{\bm{\widetilde{L}}} t\bm\right) c_2.$$ Let us define $\bm{\Theta} = c \sqrt{\bm{\widetilde{L}}}$ for convenience. In order to solve the given SPDE we need to find the inhomogeneous part. It can be found by variation of the parameters. The Wronskian of the basis functions (here denoted as $\bm{\mathbf{Wr}}$ to distinguish it from the Wiener process) is:
    \begin{equation}
        \bm{\mathbf{Wr}}(v_1, v_2)(t) = \begin{vmatrix} \cos\left(\bm{\Theta}  t\right) & \sin\left(\bm{\Theta}  t\right) \\ \frac{\mathrm{d}}{\mathrm{d}t} \cos\left(\bm{\Theta} t\right) & \frac{\mathrm{d}}{\mathrm{d}t} \sin\left(\bm{\Theta} t\right) \end{vmatrix} = \bm{\Theta},
    \end{equation}
for $\bm{v}_1 = \cos(\bm{\Theta} t)$ and $\bm{v}_2 = \sin(\bm{\Theta} t)$. The particular solution will have the form:
\begin{align} \label{equation:appendix_wave_spde_integral_form}
    \bm{u}(t) &=  \bm{\mathbf{Wr}}(\bm{v}_1, \bm{v}_2)^{-1} \left(-\bm{v}_1(t) \int {\bm{v}_2(t)}\,\mathrm{d} \bm{W}_t + \bm{v}_2(t) \int{\bm{v}_1(t)}\,\mathrm{d} \bm{W}_t\right) \\
    &= \bm{\Theta}^{-1}\left(-\cos(\bm{\Theta} t) \int_0^t \sin(\bm{\Theta} \xi)\,\mathrm{d}\bm{W}_{\xi} + \sin(\bm{\Theta} t) \int_0^t cos(\bm{\Theta} \xi)\,\mathrm{d}\bm{W}_{\xi}\right)
\end{align}

Then the solutions for the stochastic wave equation on graphs are
\begin{align}
    \bm{u}(t) &= \bm{v}_1(t) + \bm{v}_2(t) + \bm{u}_0(t) \\
    &= \cos(\bm{\Theta} t) c_1 + \sin(\bm{\Theta} t) c_2 + \bm{\Theta}^{-1}\left(-\cos(\bm{\Theta} t) \int_0^t \sin(\bm{\Theta} \xi)\,\mathrm{d} \bm{W}_{\xi} + \sin(\bm{\Theta} t) \int_0^t \cos(\bm{\Theta} \xi)\,\mathrm{d} \bm{W}_{\xi}\right).
\end{align}

Assuming that $\bm{u}(t) \sim \mathcal{N}(\bm{\mu}, Cov[\bm{u}(s), \bm{u}(t)]$, let us calculate mean and covariance from Equation~\eqref{equation:appendix_wave_spde_integral_form}:
\begin{align}
    \EX[\bm{u}(t)] &= c_1 \cos(\bm{\Theta} t)  + c_2 \sin(\bm{\Theta} t), \\
    \Cov [\bm{u}(s), \bm{u}(t)] &= \EX\left[\left(\bm{u}(s) - \EX[\bm{u}(s)]\right) \left(\bm{u}(t) - \EX[\bm{u}(t)]\right)^{\top} \right] \nonumber \\ &= \sigma^2 \bm{\Theta}^{-2} \bigg(-\cos(\bm{\Theta} t) \int_0^t \sin(\bm{\Theta} \xi)\,\mathrm{d}\bm{W}_{\xi} + \sin(\bm{\Theta} t) \int_0^t \cos(\bm{\Theta} \xi)\,\mathrm{d} \bm{W}_{\xi}\bigg) \nonumber \\
    &\qquad\quad~ \times\bigg(-\cos(\bm{\Theta} s) \int_0^s \sin(\bm{\Theta} \xi)\,\mathrm{d}\bm{W}_{\xi} + \sin(\bm{\Theta} s) \int_0^s \cos(\bm{\Theta} \xi)\,\mathrm{d}\bm{W}_{\xi}\bigg)^{\top} \nonumber \\
    &= \sigma^2 \bm{\Theta}^{-2} \bm{C}(s, t),
\end{align}
where $\bm{C}(s, t)$ is defined as
\begin{multline}
    \bm{C}(s, t) \triangleq \bigg(-\cos(\bm{\Theta} t) \int_0^t \sin(\bm{\Theta} \xi)\,\mathrm{d}\bm{W}_{\xi} + \sin(\bm{\Theta} t) \int_0^t \cos(\bm{\Theta} \xi)\,\mathrm{d} \bm{W}_{\xi}\bigg)  \\
    \times\bigg(-\cos(\bm{\Theta} s) \int_0^s \sin(\bm{\Theta} \xi)\,\mathrm{d}\bm{W}_{\xi} + \sin(\bm{\Theta} s) \int_0^s \cos(\bm{\Theta} \xi)\,\mathrm{d}\bm{W}_{\xi}\bigg)^{\top}.
\end{multline}

Opening the brackets in $\bm{C}(s, t)$ and using It\^o isometry, we will get a sum of the four following expressions:
\begin{enumerate}
    \item 
\begin{equation}
\cos(\bm{\Theta} t) \cos(\bm{\Theta} s) \int_0^{\min(t, s)} \sin^2(\bm{\Theta} \xi)\,\mathrm{d}\xi =
    \cos(\bm{\Theta} t) \cos(\bm{\Theta} s) \bigg(\frac{\xi}{2} \bm{I} - \frac{\sin(2 \bm{\Theta} \xi)}{4} \bm{\Theta}^{-1}\bigg)\Biggr|_{0}^{\min(t, s)},
\end{equation}
\item
\begin{equation}
    \sin(\bm{\Theta} t) \sin(\bm{\Theta} s) \int_0^{\min(t, s)} \cos^2(\bm{\Theta} \xi)\mathrm{d}\xi  = \sin(\bm{\Theta} t) \sin(\bm{\Theta} s) \bigg(\frac{\xi}{2} \bm{I} 
    +\frac{\sin(2 \bm{\Theta} \xi)}{4} \bm{\Theta}^{-1}\bigg)\Biggr|_{0}^{min(t, s)},
\end{equation}
\item
\begin{equation}
    - \cos(\bm{\Theta} t) \sin(\bm{\Theta} s) \int_0^{\min(t, s)} \cos(\bm{\Theta} \xi) \sin(\bm{\Theta} \xi) d\xi = -\cos(\bm{\Theta} t) \sin(\bm{\Theta} s) \bigg(-\frac{\cos^2(\bm{\Theta} \xi)}{2} \bm{\Theta}^{-1}\bigg)\Biggr|_{0}^{min(t, s)},
\end{equation}
\item
\begin{equation}
    -\sin(\bm{\Theta} t) \cos(\bm{\Theta} s) \int_0^{min(t,s)} \cos(\bm{\Theta} \xi) \sin(\bm{\Theta} \xi) d\xi = - \sin(\bm{\Theta} t) \cos(\bm{\Theta} s) \bigg(\frac{-\cos^2(\bm{\Theta} \xi)}{2} \bm{\Theta}^{-1}\bigg) \Biggr|_{0}^{min(t, s)}.
\end{equation}
\end{enumerate}

Then, the covariance is 
\begin{align}
    \bm{C}(s, t) &= \frac{\min(t, s)}{2} \left( \cos(\bm{\Theta}t) \cos(\bm{\Theta}s) + \sin(\bm{\Theta} t) \sin(\bm{\Theta}s) \right) \nonumber \\
    &\quad + \Bigg(-\cos(\bm{\Theta} t) \cos(\bm{\Theta} s) \frac{\sin(2\bm{\Theta} \min(t, s))}{4} + \sin(\bm{\Theta} t) \sin(\bm{\Theta} s) \frac{\sin(2\bm{\Theta} \min(t, s))}{4} \nonumber \\ 
    & \qquad + \cos(\bm{\Theta} t) \sin(\bm{\Theta} s) \bigg(\frac{\cos^{2}(\bm{\Theta} \min(t, s))}{2} - \frac{1}{2} \bigg) + \sin(\bm{\Theta} t) \cos(\bm{\Theta} s) \bigg(\frac{\cos^{2}(\bm{\Theta} \min(t, s))}{2} - \frac{1}{2} \bigg) \Bigg) \bm{\Theta^{-1}} \nonumber \\ 
    &= 
    \cos(\bm{\Theta} (t - s)) \, \min(t, s) - \frac{1}{2} \cos(\bm{\Theta} \max(t, s)) \sin(\bm{\Theta} \min(t, s)) \bm{\Theta}^{-1}.
\end{align}
\end{proof}

\section{Kernel Visualizations}
\label{section:appendix_kernel_visualizations}

\begin{figure*}[htbp!]
\vspace{-1em}
\hspace{1cm}\begin{adjustbox}{trim=0 0 0 2.5cm}{\begin{tikzpicture}[scale=0.92]
    \useasboundingbox (0,0) rectangle (0,4);
    \node[shape=circle,draw=black!70, line width=2pt] (A) at (0, 6) {$f_1(t)$};
    \node[shape=circle,draw=black!70, line width=2pt] (B) at (0, 4) {$f_2(t)$};
    \node[shape=circle,draw=black!70, line width=2pt] (C) at (0, 2) {$f_3(t)$};
    \path [-,line width=2pt,draw=black!70](A) edge (B);
    \path [-,line width=2pt,draw=black!70](B) edge (C);
\end{tikzpicture}\vspace{6cm}}\end{adjustbox}\vspace{-1cm}%
 \includegraphics[width=0.5\textwidth]{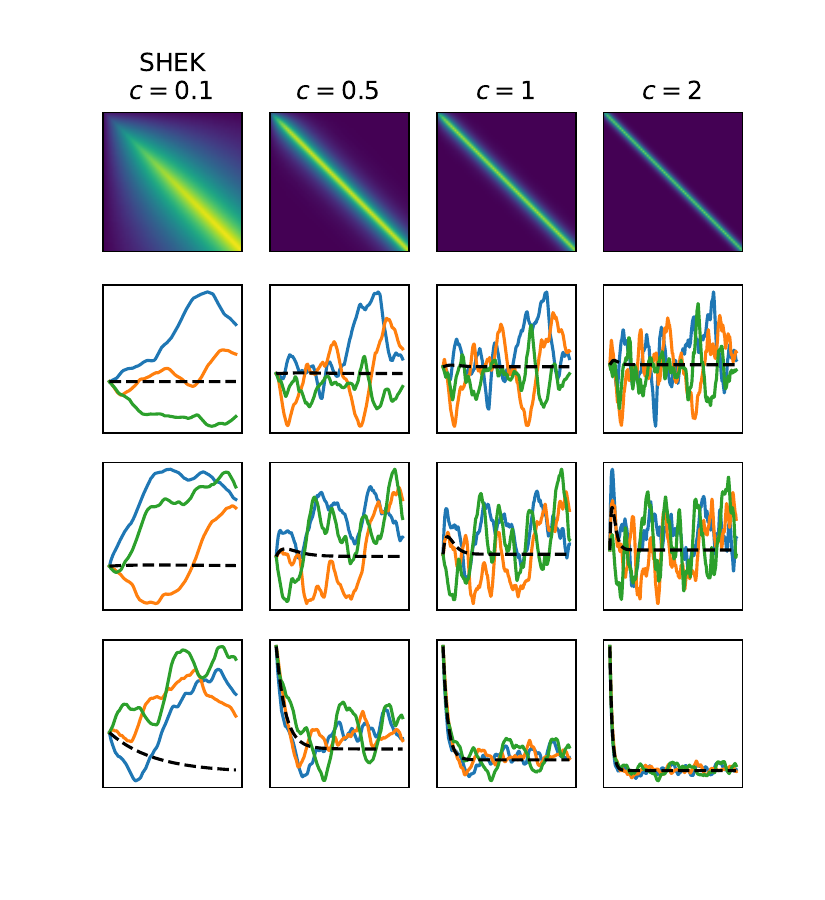}%
 \hspace{-1cm}\includegraphics[width=0.5\textwidth]{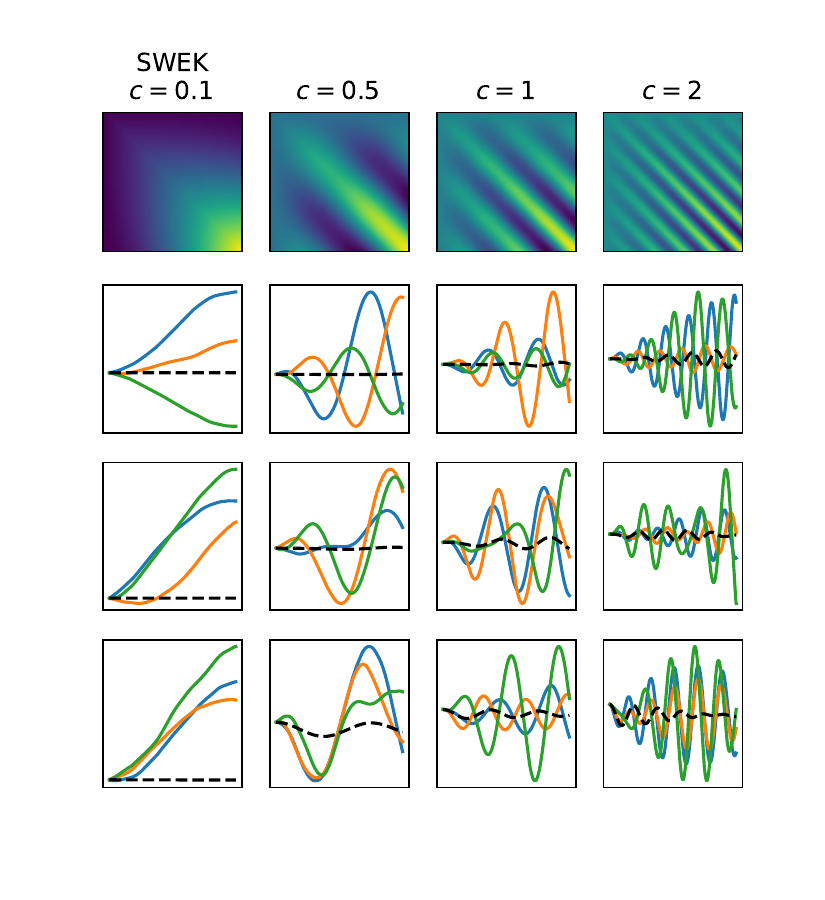}
     \caption{Temporal visualizations of SHEK (heat, left) and SWEK (wave, right) on a linear three-node graph. The first row shows the temporal part of the covariance matrix (summed over the graph vertices at each timepoint). The following three rows show mean (black) and samples (colored lines) as a function of time at each of the nodes, conditioned on $y(t=0) = (0, 0, 10)$, for different values of the hyperparameter $c$.}
    \label{figure:appendix_graph_visualizations}
\end{figure*}
\begin{figure*}[htbp!]
\vspace{-1em}
\hspace{1cm}\begin{adjustbox}{trim=0 0 0 2.5cm}{\begin{tikzpicture}[scale=0.92]
    \useasboundingbox (0,0) rectangle (0,4);
    \node[shape=circle,draw=black!70, line width=2pt] (A) at (0, 6) {$f_1(t)$};
    \node[shape=circle,draw=black!70, line width=2pt] (B) at (0, 4) {$f_2(t)$};
    \node[shape=circle,draw=black!70, line width=2pt] (C) at (0, 2) {$f_3(t)$};
    \path [-,line width=2pt,draw=black!70](A) edge (B);
    \path [-,line width=2pt,draw=black!70](B) edge (C);
\end{tikzpicture}\vspace{6cm}}\end{adjustbox}\vspace{-1cm}%
 \includegraphics[width=0.5\textwidth]{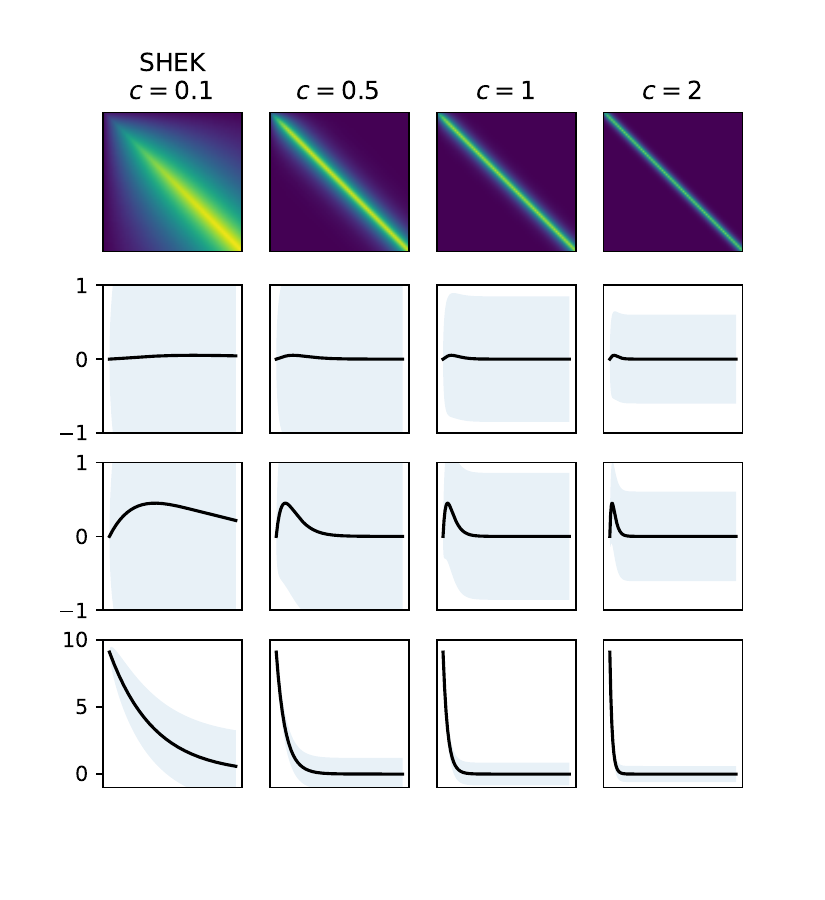}%
 \hspace{-1cm}\includegraphics[width=0.5\textwidth]{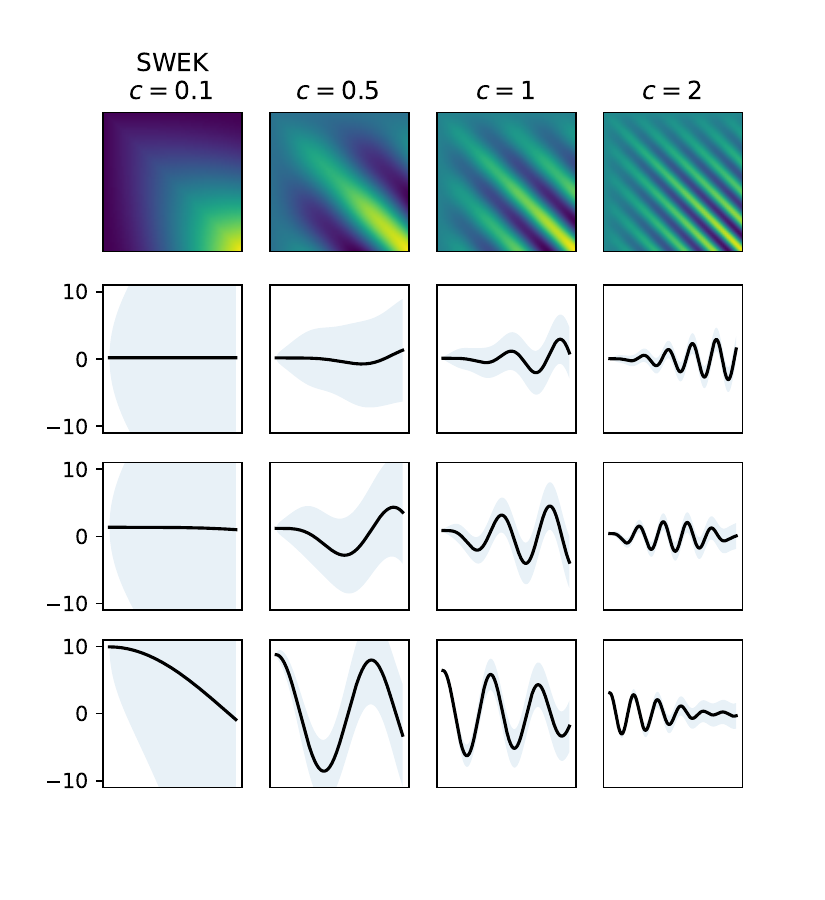}
     \caption{As in \cref{figure:appendix_graph_visualizations}, here we illustrate mean and marginal variance.}
    \label{figure:appendix_graph_visualizations2}
\end{figure*}
In Figure~\ref{figure:appendix_graph_visualizations}, we can observe behavior expected for heat and wave processes that start from the third node on a line graph. We observed similar visualizations for other types of graphs.

\clearpage
\section{Experiment Details}
\label{section:appendix_experiments_details}

\begin{figure}[htbp!]
    \begin{subfigure}[b]{0.47\textwidth}
      \begin{center}
        \includegraphics[width=\linewidth]{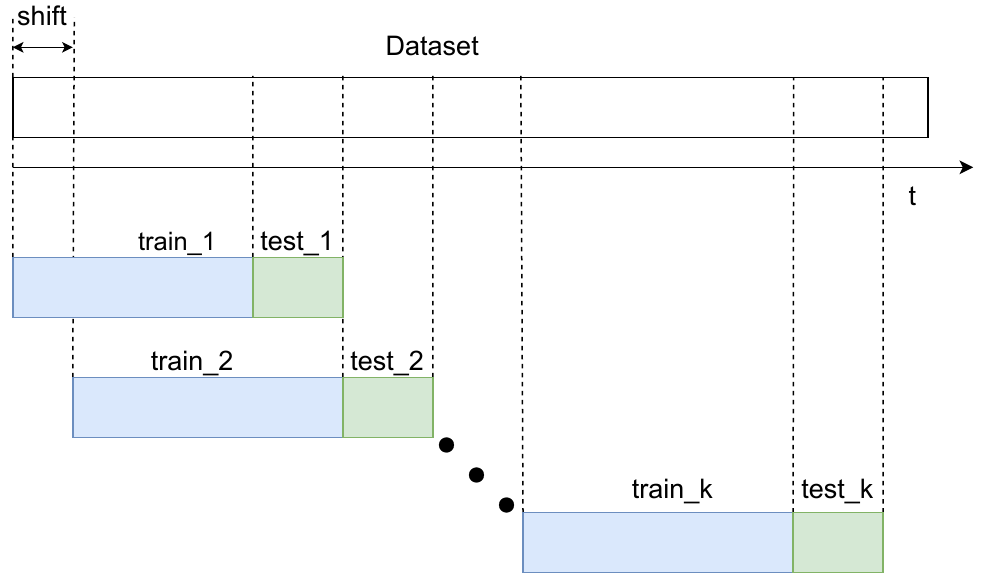}
      \end{center}
      \caption{Sliding window backtesting visualization. At each step, we select a training time interval
      (\tikz\draw[blue,fill={rgb,255:red,232; green,232; blue,252}] (0,0) rectangle (2ex, 1ex);) followed by a testing time interval (\tikz\draw[green,fill={rgb,255:red,213; green,232; blue,212}] (0,0) rectangle (1ex, 1ex);) and evaluate the performance. During the next iteration, we select time intervals that are shifted by a particular value.}
      \label{fig:appendix_sliding_window_backtest}
    \end{subfigure}
    \hfill
    \begin{subfigure}[b]{0.47\textwidth}
      \begin{center}
        \includegraphics[width=\linewidth,trim=55 40 45 40,clip]{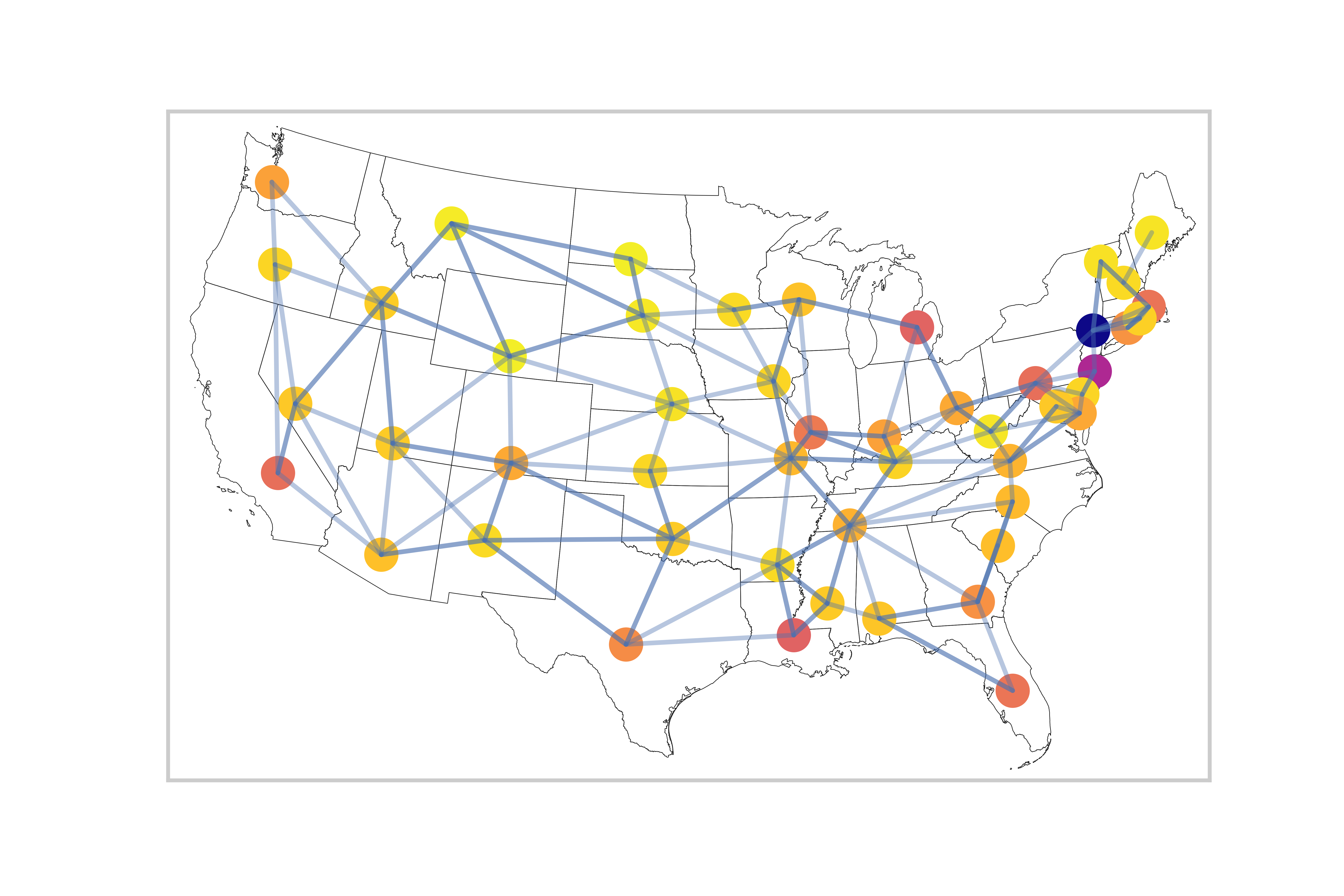}
      \end{center}
      \caption{Visualization of the graph of adjacent states. Each node is a state in the US, and each edge indicates the adjacency between two states.}
      \label{fig:appendix_visualize_covid_graph}
    \end{subfigure}
    \caption{}
\end{figure}

We ran the experiments on a NVIDIA Tesla P100-PCIE-16GB GPU. We repeated the evaluations for several validation rounds (from eight to 12 depending on the experiment) using sliding window backtesting. Sliding window backtesting is schematically visualized in Figure~\ref{fig:appendix_sliding_window_backtest} and explained in the caption. For all measurements, we report 95\% confidence intervals.

\subsection{Synthetic Wave Experiments}
\begin{wraptable}{r}{6.5cm}
\vspace{-1em}
\caption{Interpolation/extrapolation performance for the wave distribution data over a line ($\text{MAE} \times 100$).}
\label{table:appendix_1d_wave}
\setlength{\tabcolsep}{5pt}  
\footnotesize
\centering
\begin{tabular}{lcc}
\toprule
Kernel & $\text{MAE}_{\text{int}}$ & $\text{MAE}_{\text{ext}}$ \\
\midrule
SHEK($\nu$=5/2) & 0.47 $\pm$ 0.14 & 2.74 $\pm$ 0.67 \\
\textbf{SWEK($\bm{\nu}$=5/2)} & \textbf{0.28} $\pm$ \textbf{0.06} & \textbf{1.89} $\pm$ \textbf{0.28} \\
\bottomrule
\end{tabular}
\vspace{-2em}
\end{wraptable}

We generated a wave distribution process over a one-dimensional line and discretized it with a graph with 11 vertices. We performed 12 iterations of sliding window backtesting on interpolation and extrapolation tasks. For the interpolation task, we used ten percent of randomly selected measurements over 52 timepoints. For the extrapolation task, we used 50 timepoints for training followed by two timepoints where we measured generalization error. We compared SHEK($\nu$=5/2) and SWEK($\nu$=5/2). We make conclusion that MAE is better for SWEK in interpolation (DM-test, $p<0.01$) and extrapolation (DM-test, $p<0.01$) tasks. The results are presented in Table~\ref{table:appendix_1d_wave}.

\begin{figure}[htbp!]
     \centering

    \begin{subfigure}[b]{0.47\textwidth}
    \centering\scriptsize
      {\includegraphics[width=\linewidth]{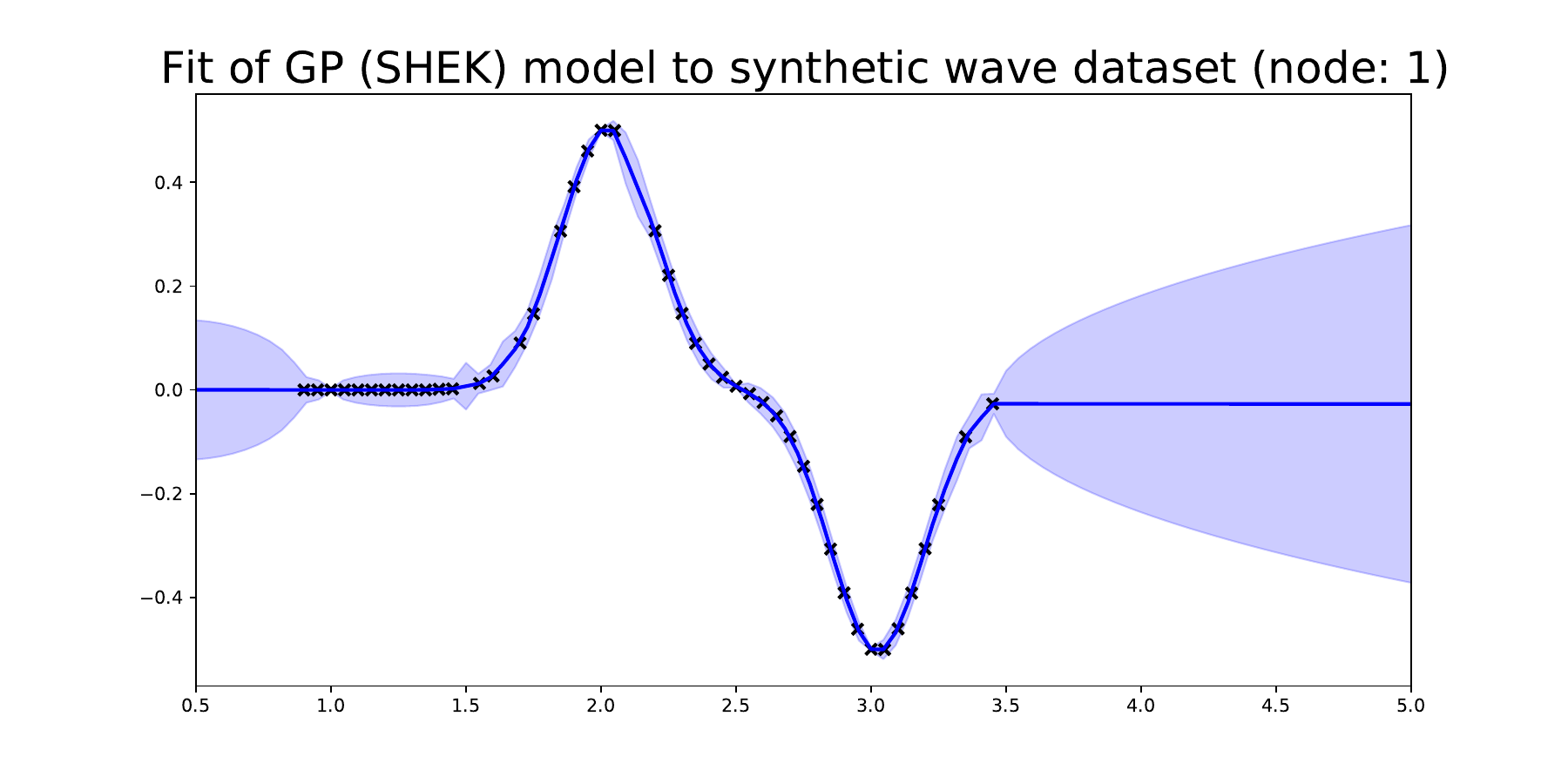}}
      \caption{SHEK on wave dataset (node \#1).}
      \label{figure:shek_1d_wave_node_1}
    \end{subfigure}
    \hfill
    \begin{subfigure}[b]{0.47\textwidth}
    \centering\scriptsize
      {\includegraphics[width=\linewidth]{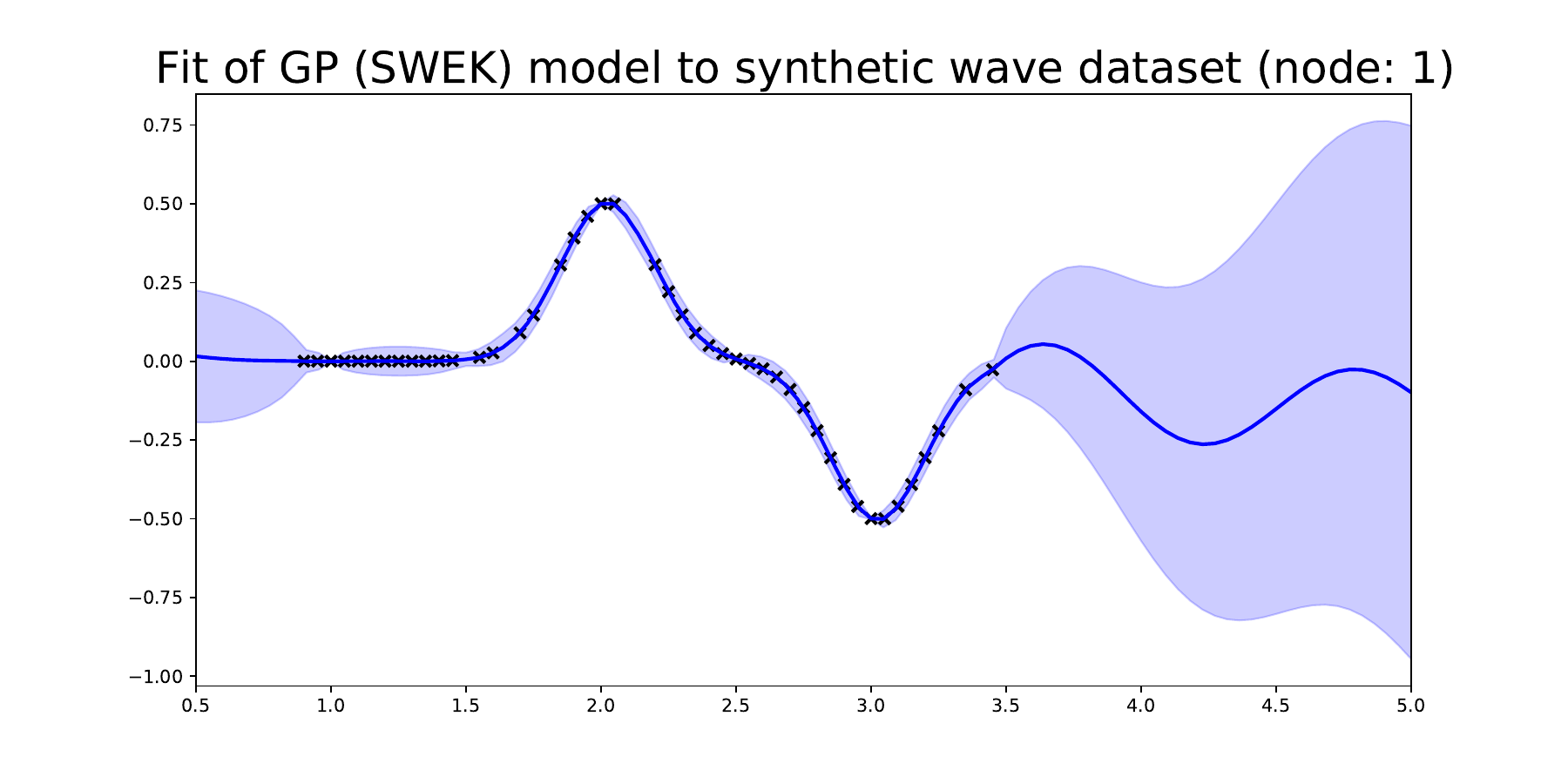}}
      \caption{SWEK on wave dataset (node \#1).}
      \label{figure:swek_1d_wave_node_1}
    \end{subfigure}
    
    \vfill
        \begin{subfigure}[b]{0.47\textwidth}
    \centering\scriptsize
      {\includegraphics[width=\linewidth]{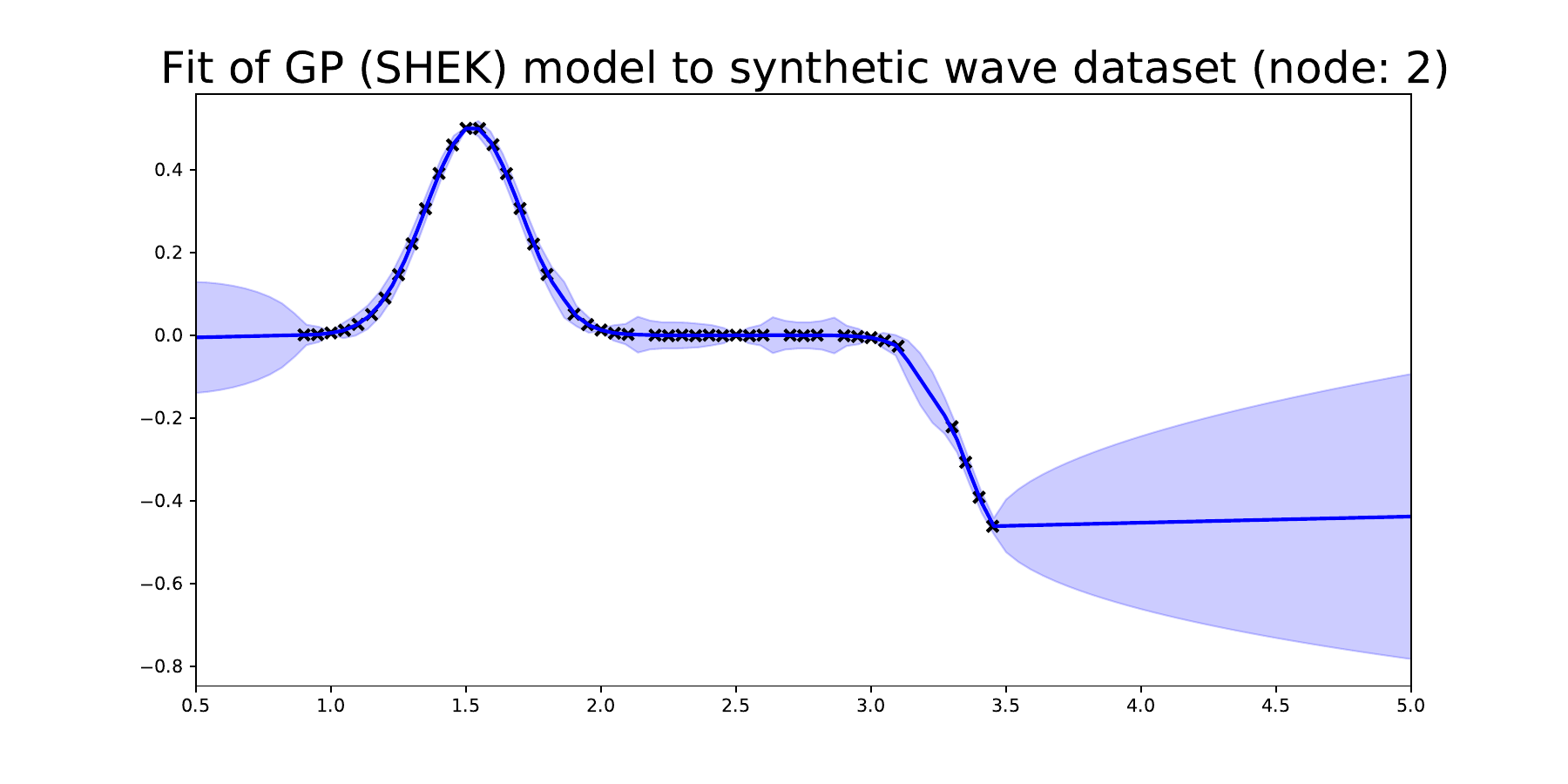}}
      \caption{SHEK on wave dataset (node \#2).}
      \label{figure:shek_1d_wave_node_2}
    \end{subfigure}
    \hfill
    \begin{subfigure}[b]{0.47\textwidth}
    \centering\scriptsize
      {\includegraphics[width=\linewidth]{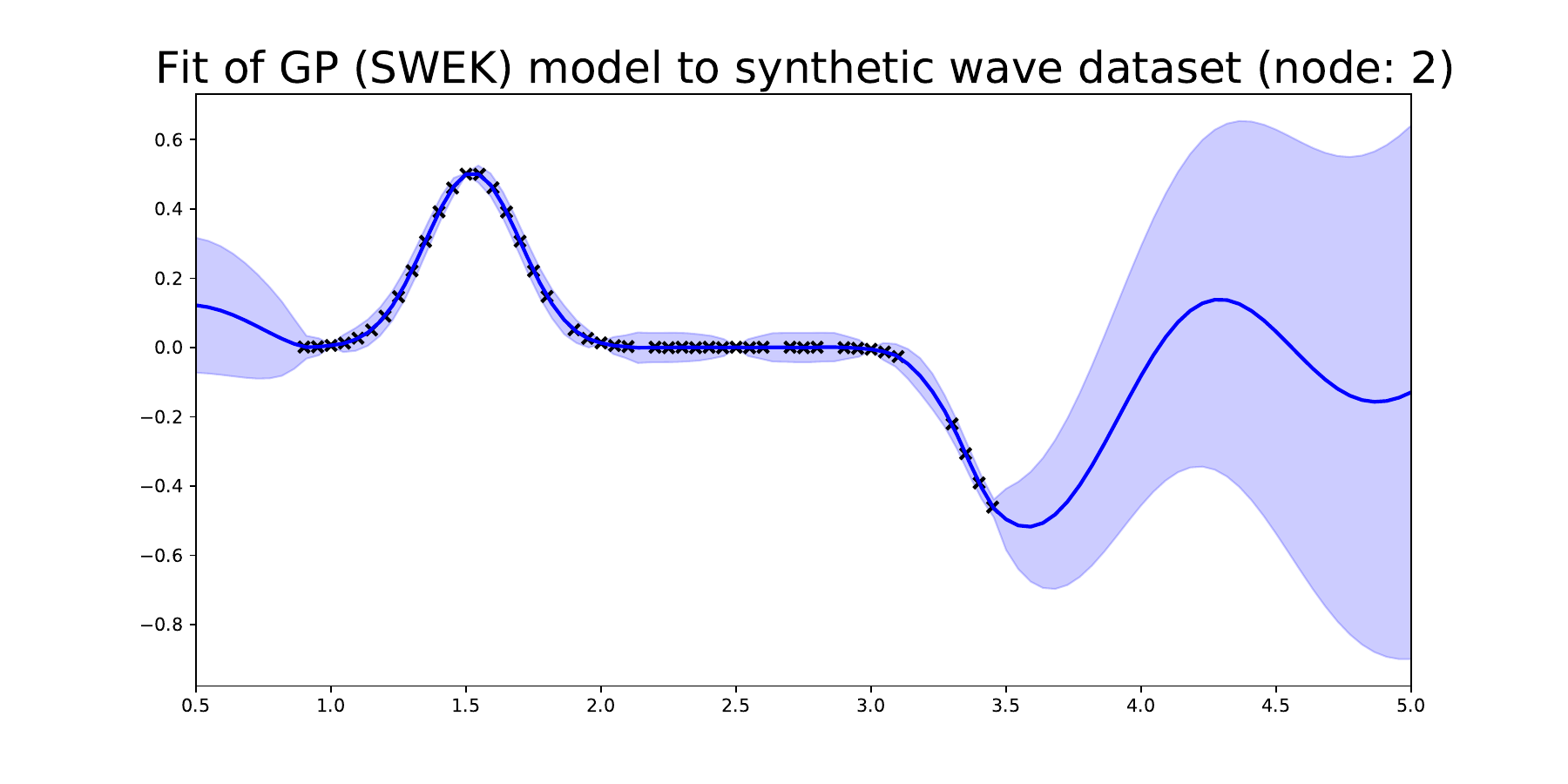}}
      \caption{SWEK on wave dataset (node \#2).}
      \label{figure:swek_1d_wave_node_2}
    \end{subfigure}

\caption{}
\label{figure:appendix_1d_wave_fit}
\end{figure}

The difference between SHEK and SWEK can be seen visually in Figure~\ref{figure:appendix_1d_wave_fit}. SWEK allows extrapolating wave behavior beyond training data, and SHEK, in contrast, generalizes badly on the synthetic wave dataset.

\subsection{Chickenpox Experiments}
\label{section:appendix_chickenpox_details}
We performed sliding window backtesting for 12 iterations.

\begin{figure}[htbp!]
     \centering

    \begin{subfigure}[b]{0.47\textwidth}
    \centering\scriptsize
      {\includegraphics[width=\linewidth]{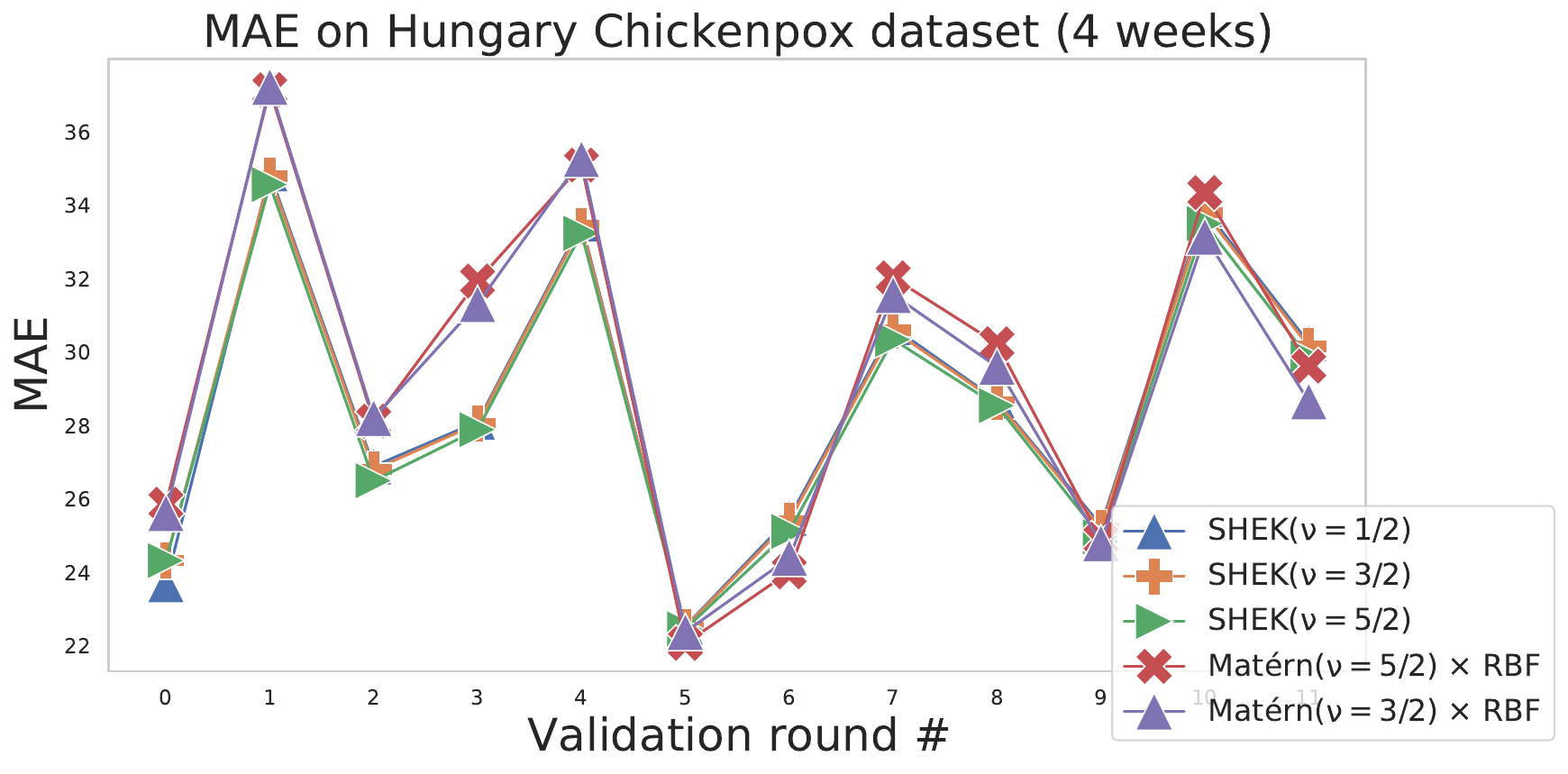}}
      \caption{Four weeks extrapolation.}
      \label{figure:chickenpox_pointwise_comparison_4}
    \end{subfigure}
    \hfill
    \begin{subfigure}[b]{0.47\textwidth}
    \centering\scriptsize
      {\includegraphics[width=\linewidth]{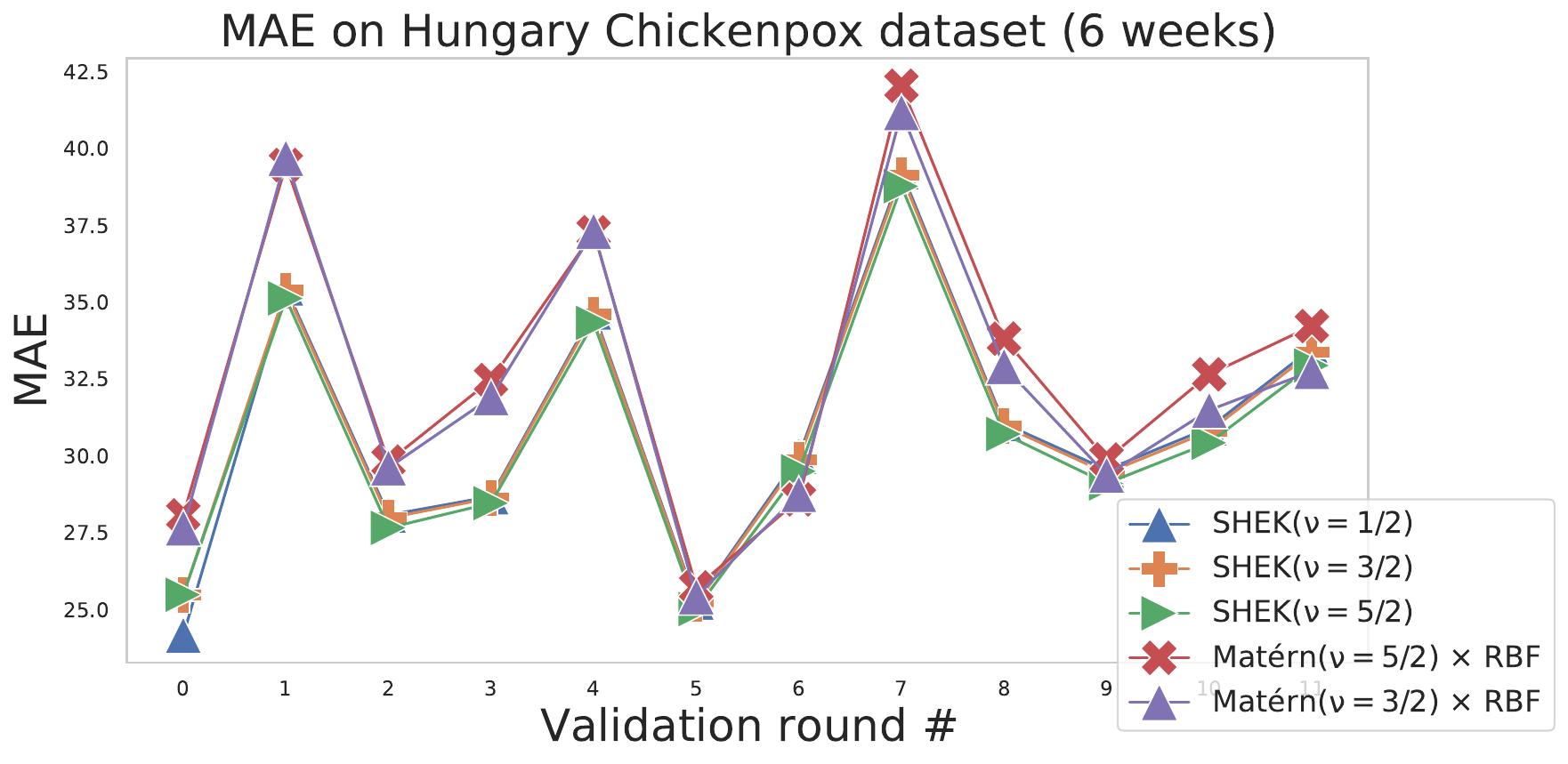}}
      \caption{Six weeks extrapolation.}
      \label{figure:chickenpox_pointwise_comparison_6}
    \end{subfigure}

\caption{Visualization of evaluation with extended extrapolation periods of four and six weeks on the Hungarian chickenpox dataset.}
\label{figure:appendix_MAE_Chickenpox}
\end{figure}

We report point-wise graphs that illustrate the comparison of generalization error on the extrapolation tasks with extended extrapolation periods in Figure~\ref{figure:appendix_MAE_Chickenpox}. The statistical significance can be measured with the Diebold-Mariano test. For example, on four weeks extrapolation period, MAE of SHEK($\nu=1/2$) is less than MAE of the product of Mat\'ern(3/2) and RBF (DM-test, $p < 0.1$), as well as on six week extrapolation period (DM-test, $p < 0.01$). 

\begin{figure}[htbp!]
     \centering

    \begin{subfigure}[b]{0.47\textwidth}
    \centering\scriptsize
      {\includegraphics[width=\linewidth]{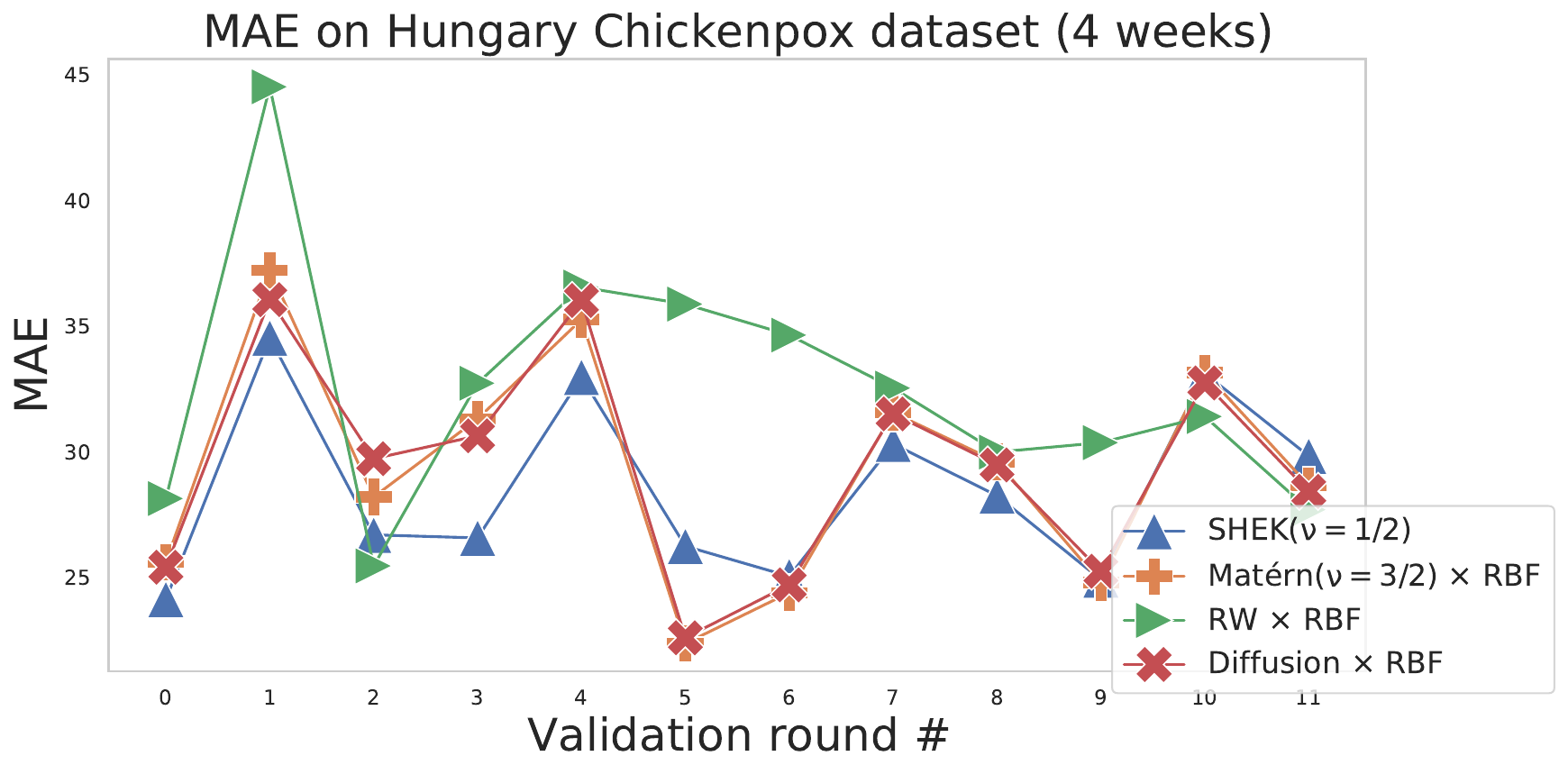}}
      \caption{Four weeks extrapolation (more separable kernels).}
      \label{figure:chickenpox_pointwise_comparison_4m}
    \end{subfigure}
    \hfill
    \begin{subfigure}[b]{0.47\textwidth}
    \centering\scriptsize
      {\includegraphics[width=\linewidth]{./images/chickenpox/mae_on_hungary_chickenpox_6_weeks.pdf}}
      \caption{Six weeks extrapolation (more separable kernels).}
      \label{figure:chickenpox_pointwise_comparison_6m}
    \end{subfigure}
\caption{Visualization of evaluation with extended extrapolation periods of four and six weeks on the Hungarian chickenpox dataset (more separable kernels).}
\label{figure:appendix_MAE_Chickenpox_more_separable}
\end{figure}

We also experimented with a larger number of separable kernels and compared them with each other and SHEK. For example, we provide results of four-weeks extrapolation in Figure~\ref{figure:appendix_MAE_Chickenpox_more_separable}.

\subsection{COVID-19 Experiments}

The dataset for this use-case consisted of two parts: information about COVID-19 cases and deaths published by The New York Times \cite{covid_dataset}, and a graph that was generated as follows. Each vertex represents a state, and two nodes $v_1$ and $v_2$ are connected if two states share a common border. The visualization of the graph is presented in Figure~\ref{fig:appendix_visualize_covid_graph}.

\begin{figure}[htbp!]
    \centering
    \begin{subfigure}[b]{0.47\textwidth}
    \centering\scriptsize
      {\includegraphics[width=\linewidth]{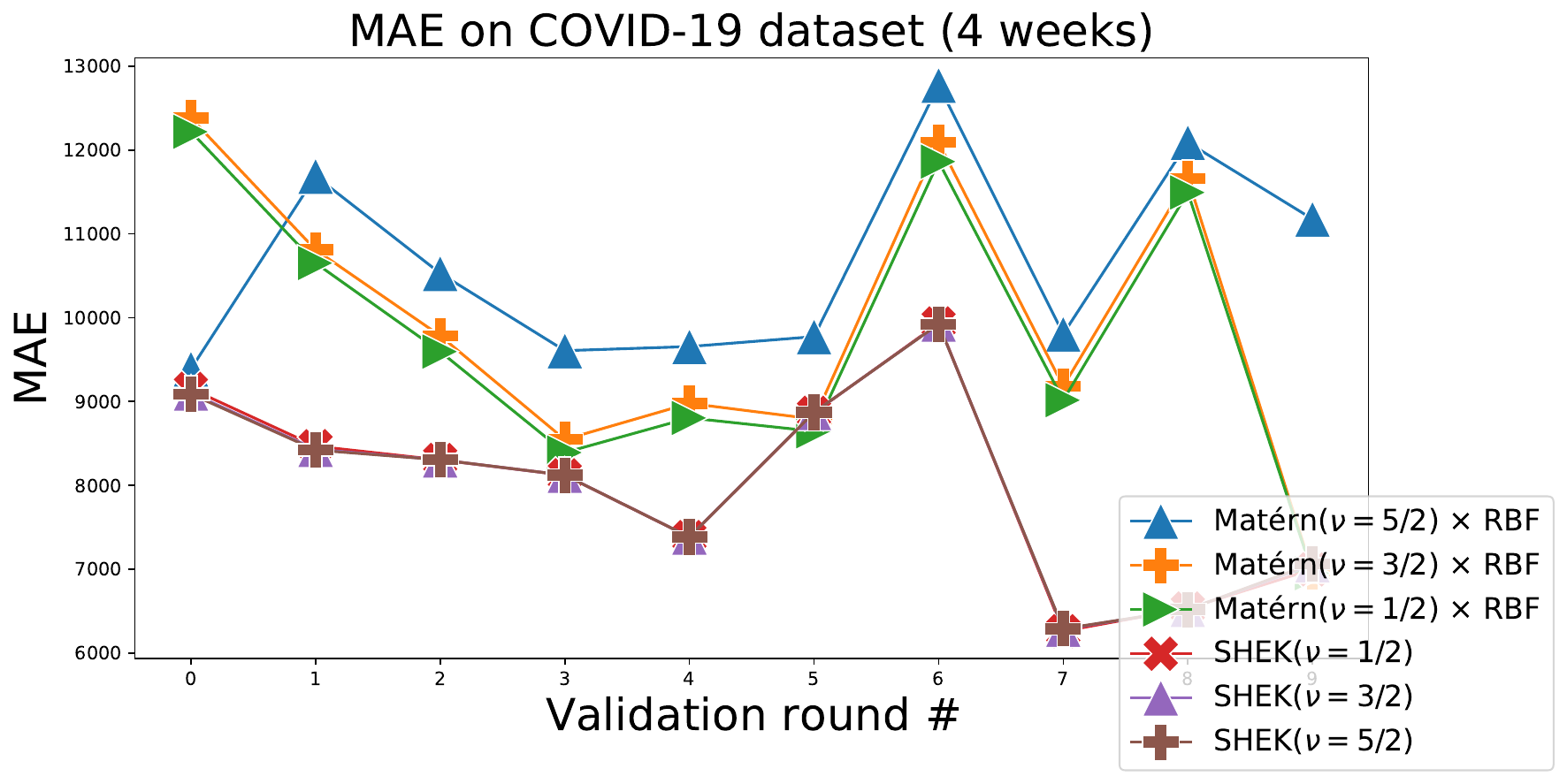}}
      \caption{Four weeks extrapolation.}
      \label{figure:covid_pointwise_comparison_4}
    \end{subfigure}
    \hfill
    \begin{subfigure}[b]{0.47\textwidth}
    \centering\scriptsize
      {\includegraphics[width=\linewidth]{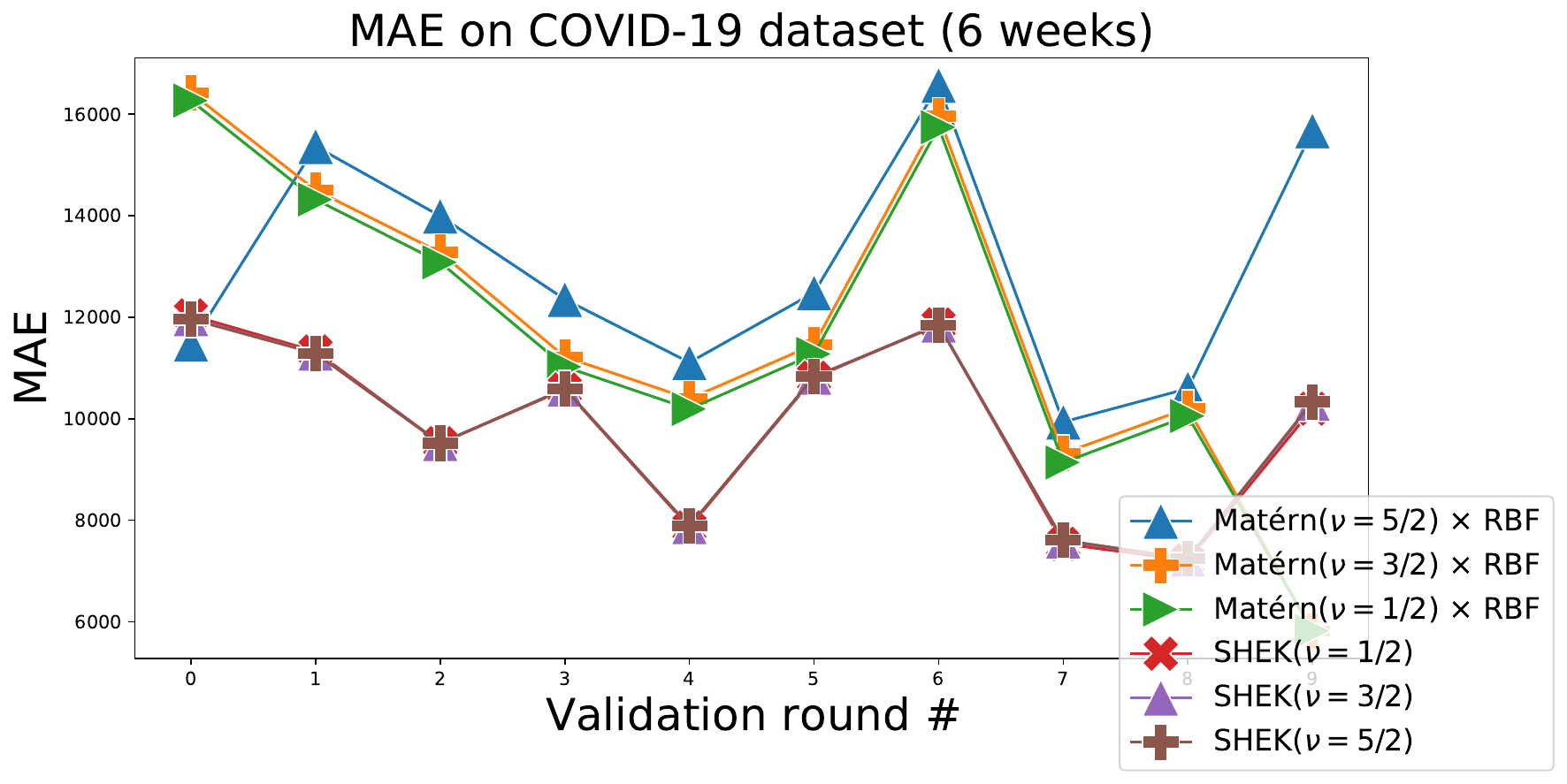}}
      \caption{Six weeks extrapolation.}
      \label{figure:covid_pointwise_comparison_6}
    \end{subfigure}

\caption{Visualization of extrapolation evaluation with extended extrapolation periods of four and six weeks on COVID-19 dataset.}
\label{figure:appendix_MAE_Covid}
\end{figure}

We evaluated the performance of the model using ten runs with a sliding window backtesting. We used 33 weeks as training data and estimated the number of cases for the following two weeks. As a metric, we used mean absolute error (MAE).

\begin{figure}[htbp!]

\begin{subfigure}[!t]{.47\linewidth}
  \includegraphics[width=\linewidth]{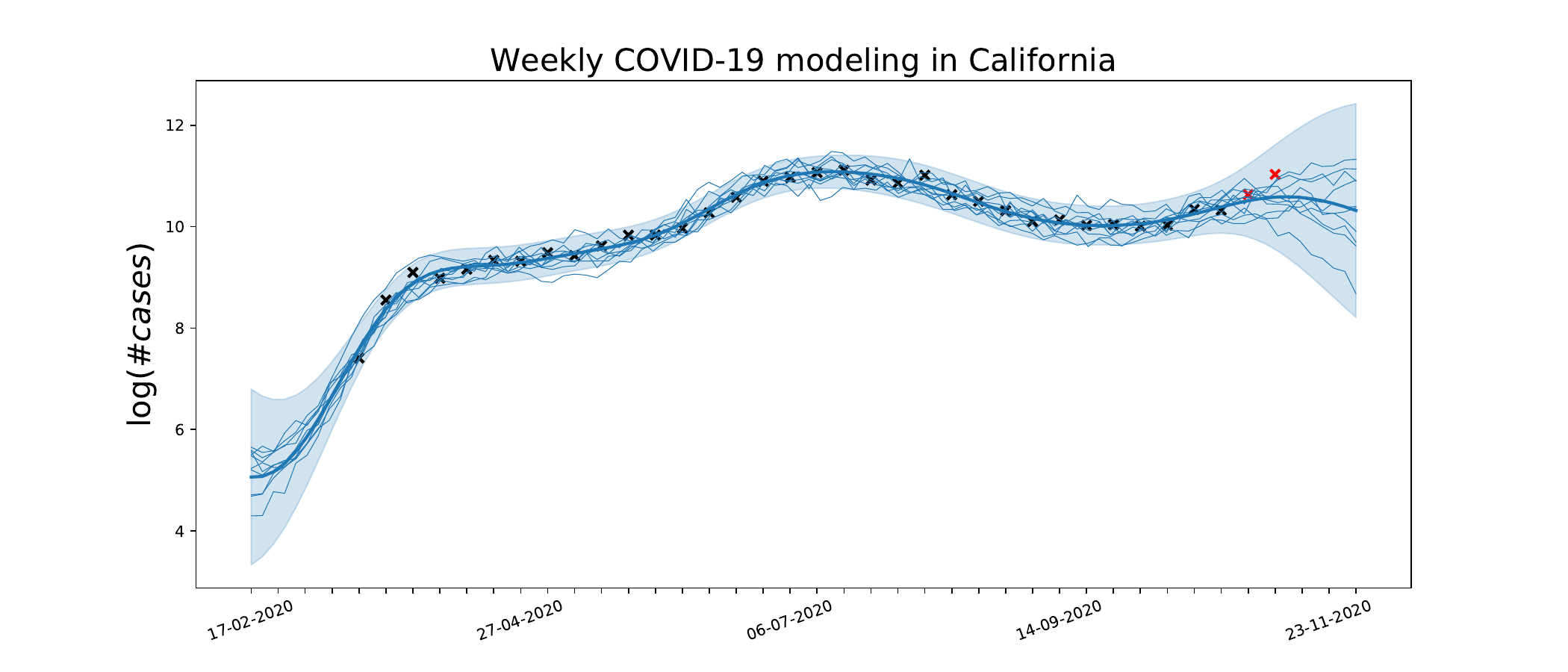}
  \caption{California}
  \label{figure:appendix_california_num_cases}
\end{subfigure}
\hfill
\begin{subfigure}[!t]{.47\linewidth}
  \includegraphics[width=\linewidth]{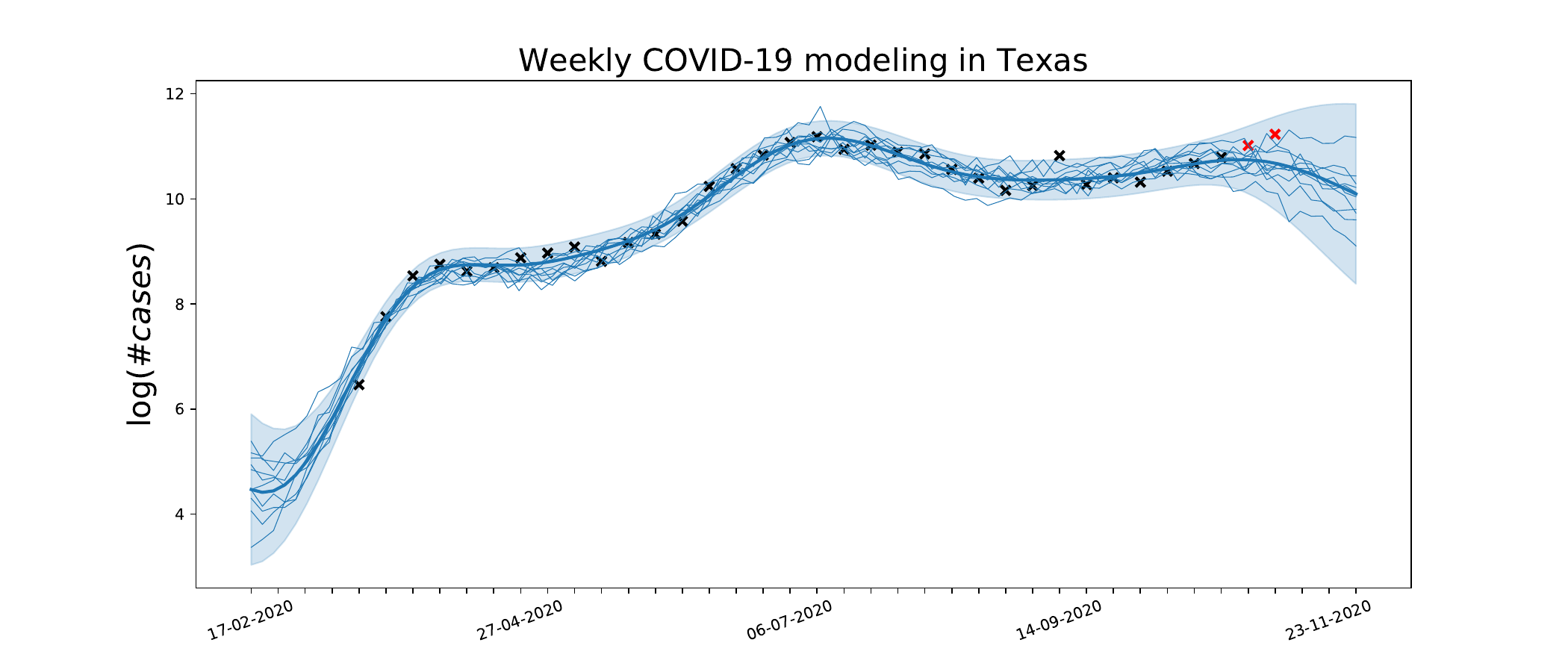}
  \caption{Texas}
  \label{figure:appendix_texas_num_cases}
\end{subfigure}
\vfill
\begin{subfigure}[!t]{.47\linewidth}
  \includegraphics[width=\linewidth]{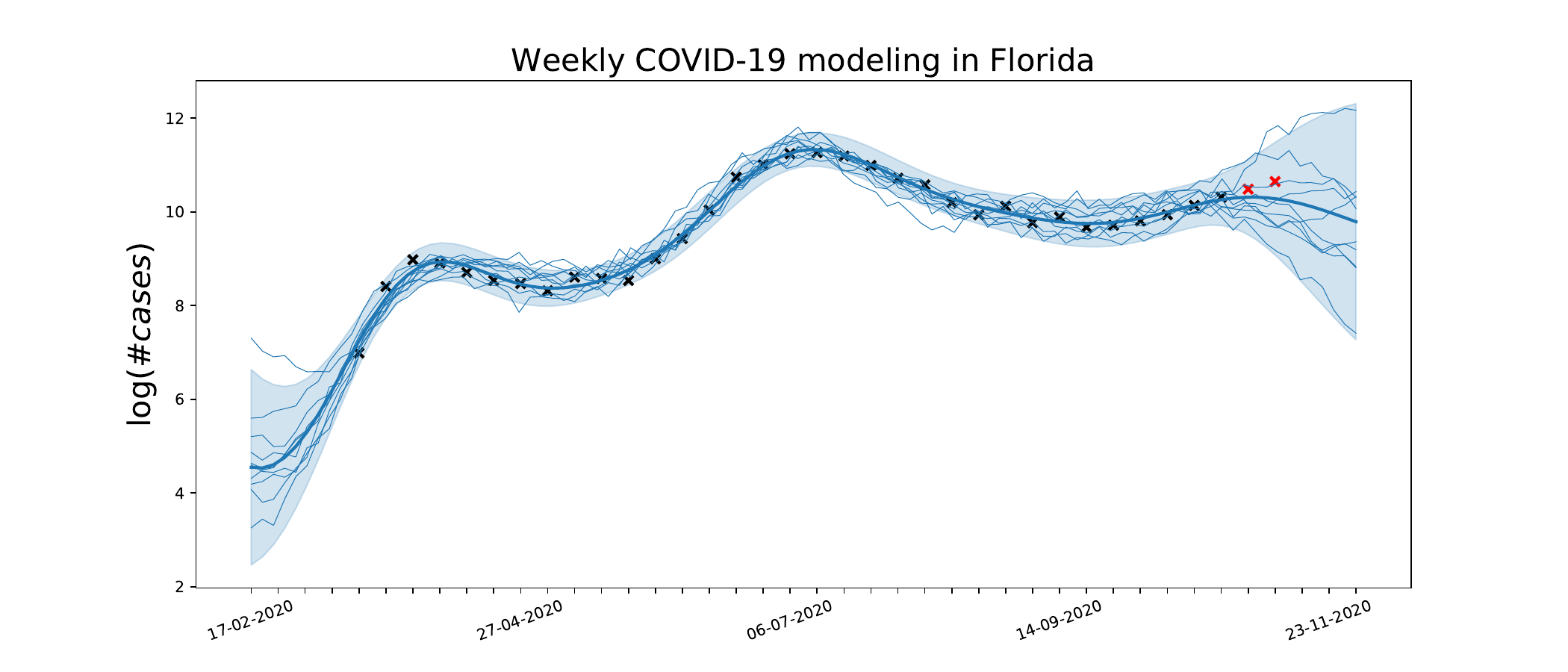}
  \caption{Florida}
  \label{figure:appendix_florida_num_cases}
\end{subfigure}
\hfill
\begin{subfigure}[!t]{0.47\linewidth}
  \includegraphics[width=\linewidth]{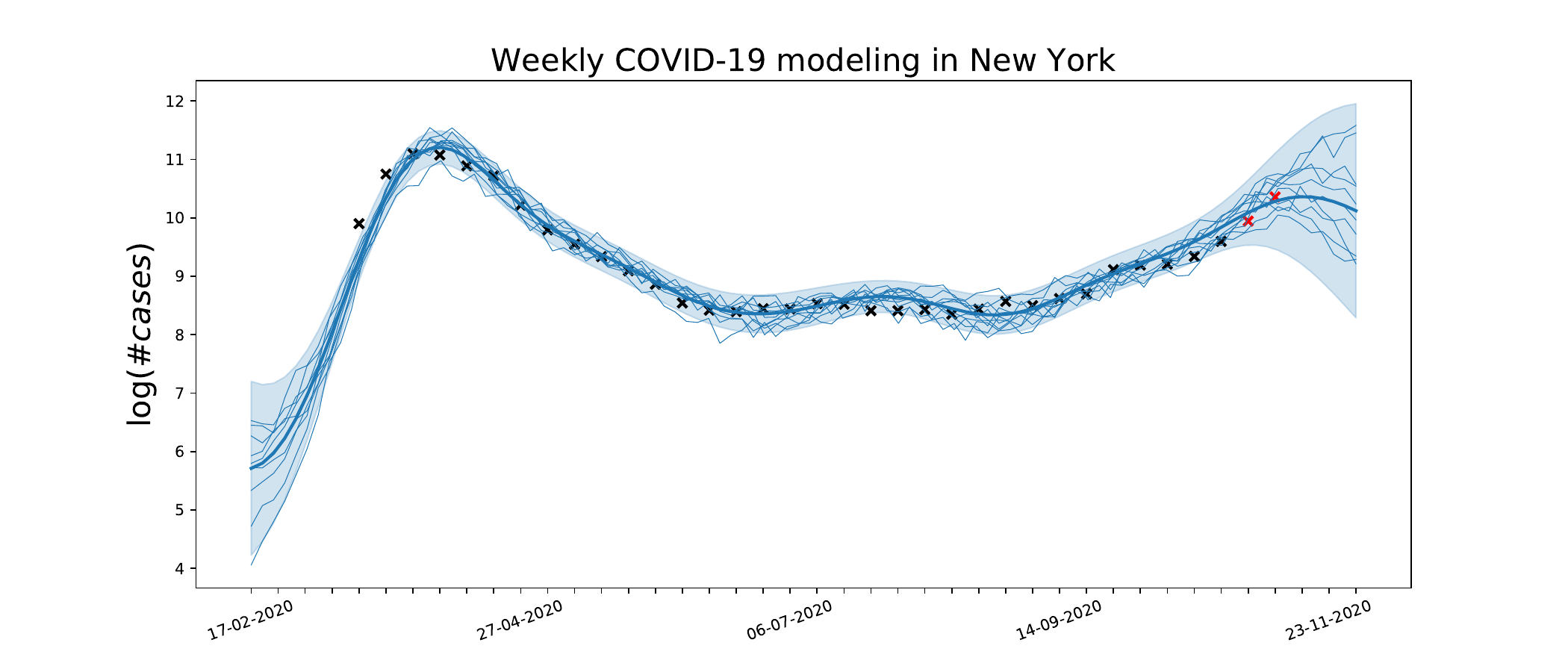}
  \caption{New York}
  \label{figure:appendix_new_york_num_cases}
\end{subfigure}

\caption{The visualizations of the GP posterior on the COVID-19 dataset for the most populated states. We indicate 97.5\% confidence intervals with blue color; we also show samples from the posterior, and training and test data points (marked with black and red crosses, respectively).}
\label{figure:appendix_num_of_cases}
\end{figure}

We additionally report the results of extrapolation over four and six weeks periods of time in Figure~\ref{figure:appendix_MAE_Covid}. We can observe that the MAE of SHEK modifications is lower than separable Mat\'ern kernels. It can be quantitatively measured using the Diebold-Mariano test. For example,  four weeks period MAE of SHEK($\nu=1/2$) is less than MAE of the product of Mat\'ern(1/2) and RBF (DM-test, $p < 0.01$), as well as on six week period (DM-test, $p < 0.05$). Next, we provide the visualizations of fitted GP using the proposed graph kernel for the three states with the largest population in the USA. In Figure~\ref{figure:appendix_num_of_cases}, we showed 97.5\% confidence intervals provided by the GP that was trained on COVID-19 dataset.

\end{document}